\documentclass[10pt,twocolumn,letterpaper]{article}

\usepackage[pagenumbers]{cvpr} 
\usepackage{multirow}
\usepackage{makecell}
\usepackage{graphicx}
\usepackage{booktabs}
\usepackage{soul}
\usepackage[table]{xcolor}
\usepackage{algorithm}
\usepackage{algorithmic}
\usepackage{caption}
\usepackage{pifont}
\usepackage[accsupp]{axessibility}

\sethlcolor{orange!30}
\definecolor{cvprblue}{rgb}{0.21,0.49,0.74}
\definecolor{myblue}{RGB}{217,225,242}
\definecolor{mygreen}{RGB}{226,239,218}
\definecolor{newgreen}{RGB}{0,176,80}
\usepackage[pagebackref,breaklinks,colorlinks,allcolors=cvprblue]{hyperref}

\def\paperID{18493} 
\def\confName{CVPR}
\def\confYear{2026}
\def\para#1{\smallskip\noindent{\bf{#1}}}

\title{Hierarchical Visual Relocalization with Nearest View Synthesis from \\Feature Gaussian Splatting}

\author{Huaqi Tao$^{1,}$\textsuperscript{*}, 
Bingxi Liu$^{1,}$\textsuperscript{*},  
Guangcheng Chen$^{1}$,   
Fulin Tang$^{2}$,  
Li He$^{1}$,  
Hong Zhang$^{1,}$\textsuperscript{\dag} \vspace{5pt} \\
$^{1}$Southern University of Science and Technology, Shenzhen, China.\\
$^{2}$Institute of Automation, Chinese Academy of Sciences, Beijing, China.\\
\tt{\small taohq2024@mail.sustech.edu.cn, hzhang@sustech.edu.cn} \\
{\small \url{https://hqitao.github.io/SplatHLoc}}
}

\begin{document}
\maketitle

\renewcommand{\thefootnote}{\fnsymbol{footnote}} 
\setcounter{footnote}{0} 
\footnotetext{\textsuperscript{*}Equal contribution. \quad \textsuperscript{\dag}Corresponding author.}

\begin{abstract}
Visual relocalization is a fundamental task in the field of 3D computer vision, estimating a camera's pose when it revisits a previously known scene.
While point-based hierarchical relocalization methods have shown strong scalability and efficiency, they are often limited by sparse image observations and weak feature matching.
In this work, we propose SplatHLoc, a novel hierarchical visual relocalization framework that uses Feature Gaussian Splatting as the scene representation.
To address the sparsity of database images, we propose an adaptive viewpoint retrieval method that synthesizes virtual candidates with viewpoints more closely aligned with the query, thereby improving the accuracy of initial pose estimation.
For feature matching, we observe that Gaussian-rendered features and those extracted directly from images exhibit different strengths across the two-stage matching process: the former performs better in the coarse stage, while the latter proves more effective in the fine stage. Therefore, we introduce a hybrid feature matching strategy, enabling more accurate and efficient pose estimation.
Extensive experiments on both indoor and outdoor datasets show that SplatHLoc enhances the robustness of visual relocalization, setting a new state-of-the-art.
\end{abstract}
\vspace{-11pt}
\section{Introduction}
\label{sec:intro}

\begin{figure}[htbp]
    \centering
    \includegraphics[width=0.45\textwidth]{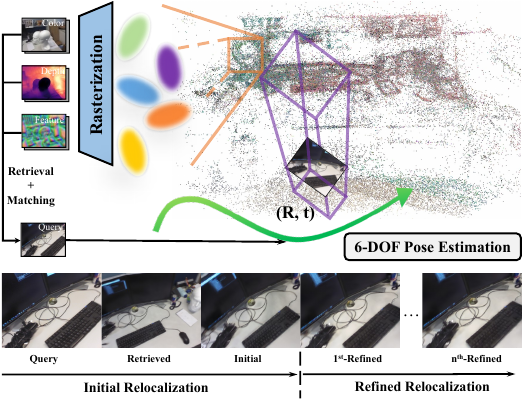}
    \caption{\textbf{SplatHLoc: a novel hierarchical visual relocalization framework based on Feature Gaussian Splatting (FGS).} FGS renders \textit{color}, \textit{depth}, and \textit{feature} maps from novel views, which our method exploits to improve the image retrieval and feature matching process. Upon retrieving a reference image, we match it to the query to estimate an initial pose (initial relocalization). We then render views from the estimated pose and iteratively match them to the query to refine the pose (refined relocalization).}
    \vspace{-11pt}
    \label{fig: intro}
\end{figure}

Visual relocalization, which seeks to estimate the 6-DoF poses of query images within a given environment \cite{piasco2018survey}, underpins various critical applications, including intelligent robotics \cite{biswas2012depth}, augmented reality \cite{ventura2014global}, and autonomous driving \cite{moreau2022coordinet}. Current approaches to visual relocalization can be broadly categorized into three groups based on their scene representations: structure-based, regression-based, and render-based methods. 

\textit{Structure-based methods} represent the environment as a sparse 3D point cloud reconstructed via Structure-from-Motion (SfM) \cite{schonberger2016structure, schoenberger2016mvs}.
The camera pose is estimated using the Perspective-n-Point (PnP) \cite{lepetit2009ep} algorithm using 2D-3D correspondences obtained either by directly matching 2D keypoints to the 3D point clouds \cite{liu2017efficient, sattler2012improving}, or through a hierarchical framework that performs image retrieval followed by feature matching \cite{sarlin2019coarse, taira2018inloc}. In particular, hierarchical approaches (e.g., HLoc \cite{sarlin2019coarse}, Kapture~\cite{humenberger2022investigating}) demonstrate superior scalability to large scenes and robustness to substantial appearance changes.
\textit{Regression-based methods} primarily leverage neural networks to encode both geometric and semantic information of a scene. Trained on posed images, these models regress either absolute camera pose \cite{kendall2015posenet, chen2022dfnet, shavit2022camera, kendall2017geometric, liu2024hr} or pixel-wise scene coordinates \cite{brachmann2018learning, brachmann2021visual, brachmann2023accelerated, tang2023neumap, revaud2024sacreg, li2020hierarchical, bruns2025ace}. 
Despite these advances, both structure-based and regression-based pipelines exhibit inherent limitations in handling sparse observations and complex scene appearance, motivating a complementary line of research that explores \textit{richer} scene representations for relocalization.

Recently, \textit{render-based methods} have gained attention by utilizing scene representations capable of Novel View Synthesis (NVS), including meshes \cite{panek2022meshloc, li2025gpvk}, Neural Radiance Fields (NeRF) \cite{pietrantoni2024self, lin2022parallel, yen2021inerf, zhou2024nerfect}, and Gaussian Splatting (GS) \cite{pietrantoni2025gaussian, botashev2024gsloc, sidorov2025gsplatloc, cheng2025logs, zhai2025splatloc}.
In particular, GS \cite{kerbl20233d, huang20242d} has emerged as a superior scene representation, offering higher-quality and more efficient rendering.
Some works leverage rendered images from novel views to augment the database with additional keyframes for image retrieval \cite{li2025gpvk}, or network training \cite{li2025adversarial}, but this often comes at the cost of increased storage needs and longer search times.
Another important research direction is leveraging rendered data for pose refinement \cite{liu2025gscpr, huang2025sparse}. 
STDLoc \cite{huang2025sparse} aligns rendered features from Feature Gaussian Splatting (FGS) \cite{zhou2024feature} with query features encoded from images for dense matching, achieving promising results.
However, we observe that this asymmetric matching is suboptimal. 
Our \textit{key insight} is that rendered features inherently provide multi-view knowledge and mitigate the accumulated errors introduced when extracting features from rendered images, but they are limited in modeling fine geometric relations due to the feature gap with query features. 
Therefore, the rendered features are more suitable for patch-level coarse matching rather than pixel-level fine matching.

In this paper, we propose SplatHLoc, a novel hierarchical visual relocalization framework built upon the FGS scene representation, as illustrated in Fig.~\ref{fig: intro}. 
We begin by retrieving the top-$k$ candidate images from the posed training set and perform geometric verification \cite{yu2024gv} to identify the one most similar to  the query.
\textit{To address the challenge of sparse image observations}, we apply random perturbations to the retrieved pose and render novel-view color images from the FGS representation.
The retrieval and geometric verification are then repeated over these synthesized views to determine a reference image that is closest to the query viewpoint.
Once the nearest view is identified, an initial pose is estimated by matching the query image with the synthesized reference, followed by iterative refinement through repeated rendering and matching.
\textit{To ensure accuracy and efficient feature matching,} we first align the rendered feature map with the coarse features extracted from the query during the coarse matching stage. 
For fine matching, we employ a well-trained semi-dense matcher to extract fine features from both the rendered and query images. Guided by the coarse correspondences, these fine-level features further refine the matches down to the pixel-level.

In summary, our contributions are as follows:

\begin{itemize}

\item We propose an adaptive viewpoint retrieval method that perturbs the initially retrieved pose to render virtual keyframes. Image retrieval over these virtual keyframes significantly improves the robustness of initial pose estimation, especially in weakly textured regions.

\item We introduce a hybrid feature matching strategy, using the rendered features for coarse matching and the fine features from a semi-dense matcher for fine matching, enhancing the accuracy and efficiency of pose estimation.

\item We conduct extensive experiments to validate the robustness and effectiveness of our proposed method, achieving new state-of-the-art results on both indoor and outdoor public benchmarks.

\end{itemize}

\section{Related Work}
\label{sec:related}

\begin{figure*}[htbp]
    \centering
    \includegraphics[width=0.98\textwidth]{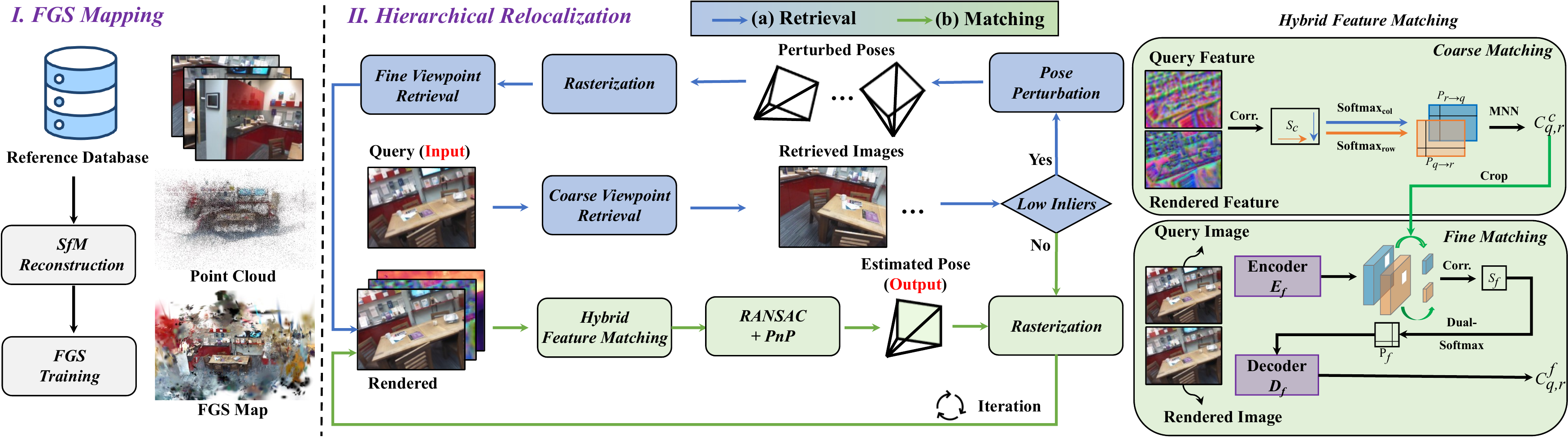}
    \caption{\textbf{An overview of the proposed SplatHLoc framework.} Starting from a database of reference images, we build an SfM model to initialize the Gaussian primitives, and then training the FGS map, see Section \ref{3.1}. SplatHLoc follows a hierarchical relocalization pipeline. (a) In the \colorbox{myblue}{retrieval} stage, we propose an adaptive coarse-to-fine viewpoint retrieval strategy. We first perform the coarse retrieval to obtain retrieved images and then use a lightweight feature matcher to perform geometric verification for each query–retrieved image pair. If geometric verification yields fewer inliers than a threshold, we perform the fine viewpoint retrieval to get a fine retrieved pose, see Section \ref{3.2}. (b) In the \colorbox{mygreen}{matching} stage, we apply the proposed hybrid feature matching strategy to establish 2D–2D correspondences between the query and the retrieved image, see Section \ref{3.3}. Next, the rendered depth map lifts the 2D–2D matches to 2D–3D. We then estimate an initial pose using RANSAC-PnP. Rendering from the estimated pose and repeating the matching stage enables pose refinement. The overall relocalization process is further discussed in Section \ref{3.4}.
    }
    \vspace{-11pt}
    \label{fig: overview}
\end{figure*}

\para{Hierarchical Visual Relocalization.}
Hierarchical relocalization leverages image retrieval, also referred to as Visual Place Recognition (VPR) \cite{lowry2015visual}, to determine a coarse location of the query, followed by image matching to estimate the fine pose.
A representative example of this pipeline is HLoc \cite{sarlin2019coarse}, which constructs a sparse 3D map using COLMAP \cite{schonberger2016structure} with 2D–3D correspondences extracted from database images. During relocalization stage, it retrieves similar images and matches features to obtain 2D–2D correspondences, which are then lifted to 2D–3D correspondences for pose estimation.
Similarly, InLoc \cite{taira2018inloc} relies on a database of posed RGB-D images captured in indoor environments. After retrieving a visually similar reference image and establishing 2D–2D correspondences with the query, depth maps are used to lift keypoints into 3D to form 2D–3D correspondences.
Owing to their modular design, these approaches can flexibly integrate different image retrieval \cite{ali2023mixvpr, izquierdo2024optimal}, feature detection \cite{detone2018superpoint, potje2024xfeat}, and feature matching \cite{sarlin2020superglue, lindenberger2023lightglue, sun2021loftr} algorithms, making them adaptable to different accuracy–efficiency trade-offs.
However, their performance fundamentally depends on the availability of database images with viewpoints sufficiently close to the query. When such images are missing, establishing reliable feature correspondences becomes challenging, ultimately degrading relocalization accuracy.
GPVK-VL \cite{li2025gpvk} attempts to address this by placing and rendering virtual keyframes prior to retrieval, thereby augmenting the database. Yet, the synthesized keyframes are not guaranteed to align with the query viewpoint and incur substantial storage overhead, limiting the scalability.

In contrast, our method renders additional views around the initially retrieved images, enabling the system to retrieve views with viewpoints are closer to that of the query. This results in more reliable feature correspondences.
Furthermore, we perform a second retrieval only when the initially retrieved image produces too few inlier matches with the query. Benefiting from the efficiency of both Gaussian rendering and modern image retrieval techniques, our approach maintains overall computational efficiency.

\para{Novel View Synthesis for Pose Refinement.} 
To further improve relocalization accuracy, several works leverage the view synthesis capability of NeRF \cite{moreau2023crossfire, zhou2024nerfect, zhao2024pnerfloc} or GS \cite{cheng2025logs, sidorov2025gsplatloc, botashev2024gsloc, huang2025sparse, liu2025gscpr} to refine the estimated query pose.
These approaches refine an initially estimated pose by rendering an image from the estimated viewpoint and aligning it with the query image, either via photometric error minimization \cite{cheng2025logs, sidorov2025gsplatloc, botashev2024gsloc} or feature matching \cite{huang2025sparse, liu2025gscpr}.
Among them, photometric error minimization suffers from sensitivity to illumination variations, which limits its reliability in real-world applications.
On the other hand, rendered images from sparse training viewpoints often exhibit floaters, which can significantly hinder accurate feature matching.
While GS-CPR \cite{liu2025gscpr} utilizes MASt3R \cite{leroy2024grounding}, a powerful foundation-model-based matcher, for pose refinement, it suffers from low input resolution and high computational demand.
STDLoc \cite{huang2025sparse} achieves promising results under low-texture and illumination variations via coarse-to-fine matching of rendered features, but it overlooks the feature gap between rendered and query features.

Unlike prior works, our method utilizes rendered features for coarse matching, and a semi-dense matchers to extract fine features that refine correspondences. This hybrid design enable accurate pose estimation, achieving better relocalization performance than previous SotA methods.

\vspace{-8pt}
\section{Method}

In this section, we detail our method. We begin with preliminaries for FGS in Section \ref{3.1}.
We then present our adaptive viewpoint retrieval method in Section \ref{3.2}.
We discuss our hybrid feature matching strategy, which combines coarse features from FGS with the fine features from a trained encoder in Section \ref{3.3}.
Finally, we describe the overall relocalization pipeline of our method in Section \ref{3.4}.
An overview of the proposed method is provided in Fig. \ref{fig: overview}.

\subsection{Preliminaries on FGS}
\label{3.1}
We adopt FGS as our scene representation, which extends 3DGS by incorporating a feature field. 
We initialize the Gaussian primitives from the SfM point cloud.
For $i$-th Gaussian primitive, it is associated with a set of trainable attributes, including its center $\mathbf{x}_i \in \mathbb{R}^3$, rotation quaternion $\mathbf{q}_i \in \mathbb{R}^4$, scale $\mathbf{s}_i \in \mathbb{R}^3$, opacity $\alpha_i \in \mathbb{R}$, color $\mathbf{c}_i \in \mathbb{R}^3$, alone with feature $\mathbf{f}_i \in \mathbb{R}^d$. Formally, each Gaussian primitive is denoted as:
$
\mathcal{G}_i = \{\mathbf{x}_i, \mathbf{q}_i, \mathbf{s}_i, \alpha_i, \mathbf{c}_i, \mathbf{f}_i\}.
$

\begin{figure}[htbp]
    \centering
    \includegraphics[width=0.46\textwidth]{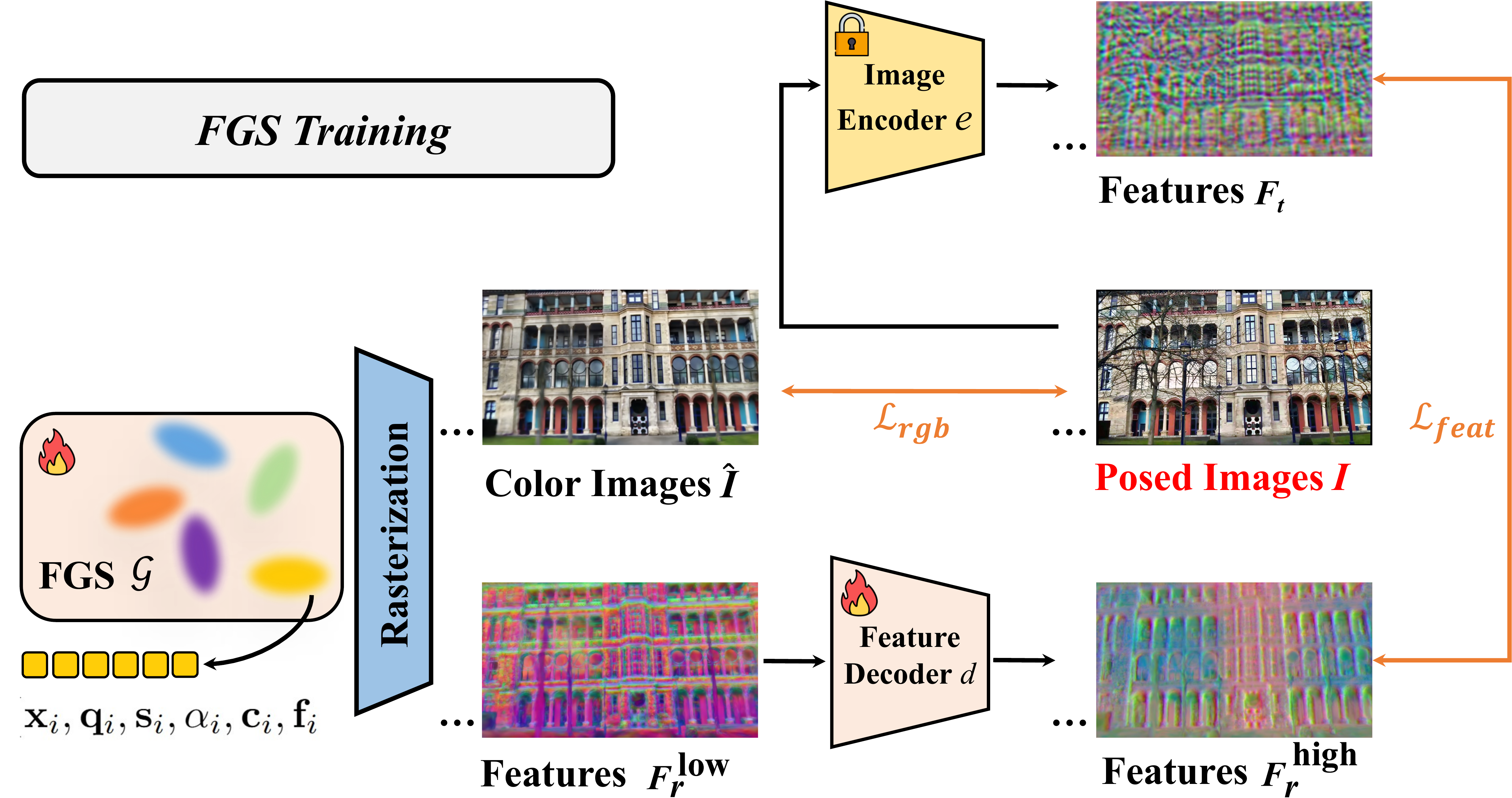}
    \caption{\textbf{Illustration of the FGS training process.} Feature decoder $d$ is introduced to reduce the dimensionality of the rendered feature $F_r^{\text{low}}$ for improved efficiency and reduced map size.}
    \vspace{-11pt}
    \label{fig: training}
\end{figure}

\para{Training Process.} 
The training process follows Feature 3DGS \cite{zhou2024feature}, jointly optimizing the radiance field and the feature field, as shown in Fig.~\ref{fig: training}. In addition, we \textit{compress} the feature field to improve feature-rendering efficiency and reduce storage overhead \cite{yue2024improving}.

Given a training image $I \in \mathbb{R}^{3 \times H \times W}$, we extract the dense feature $F_t$ using an image encoder $e$: 
\begin{equation}
F_t = \text{Resize}(e(I)), \quad F_t \in \mathbb{R}^{C \times H' \times W'}.
\end{equation}
The FGS render the color image $\hat{I}\in \mathbb{R}^{3 \times H \times W}$ and the feature map $F_r^{\text{low}} \in \mathbb{R}^{C' \times H \times W}$ from the training viewpoint, where $C' \ll C$. To restore the full feature dimensionality, $F_r^{\text{low}}$ is further passed through a $3 \times 3$ convolution that serves as a scene-specific feature decoder $d$ to increase its channel dimension.
The resulting feature map is then resized via bilinear interpolation to align with $F_t$:
\begin{equation}
F_r^{\text{high}} = \text{Resize}(d(F_r^{\text{low}})), \quad F_r^{\text{high}} \in \mathbb{R}^{C \times H' \times W'}.
\end{equation}

We optimize the parameters of the FGS and the feature decoder $d$ by minimizing a weighted sum of photometric loss $\mathcal{L}_{\text{rgb}}$ and feature loss $\mathcal{L}_{\text{feat}}$:
\begin{equation}
\label{loss}
\mathcal{L} = \mathcal{L}_{\text{rgb}} + \gamma \mathcal{L}_{\text{feat}}, 
\end{equation}
where $\mathcal{L}_{\text{rgb}}$ combines $\mathcal{L}_1$ and D-SSIM losses between the ground-truth image $I$ and the rendered image $\hat{I}$, and $\mathcal{L}_{\text{feat}}$ is an $\mathcal{L}_1$ loss between the ground-truth feature map $F_t$ and the rendered feature map $F_r^{\text{high}}$:
\vspace{-3pt}
\begin{align}
\mathcal{L}_{\text{rgb}} &= (1 - \lambda)\, \mathcal{L}_1(I, \hat{I}) + \lambda\, \mathcal{L}_{\text{D-SSIM}}(I, \hat{I}), \label{eq:Lrgb}\\
\mathcal{L}_{\text{feat}} &= \mathcal{L}_1\left( F_t, F_r^{\text{high}} \right).  \label{eq:Lfeat}
\end{align}

\subsection{Adaptive Coarse-to-Fine Viewpoint Retrieval}
\label{3.2}

For scalability and efficiency of initial relocalization, we employ image retrieval to find reference images captured from the same place as the query.
However, the reference images are \textit{unevenly distributed} within the scene, and some regions suffer from \textit{sparse observations}. As a result, the retrieved images may exhibit significant viewpoint differences from the query. 
To address this, our method adopts a coarse-to-fine (C2F) image retrieval pipeline. The conventional retrieval process serves as the coarse stage, while the fine stage leverages the FGS map to synthesize novel views that are closer to the query viewpoint.

\para{Coarse Viewpoint Retrieval.} 
We first aggregate all training and query images into compact global descriptors using a VPR method $\mathcal{V}$. 
Then, for each query image $I_q$, we retrieve the top-$k_1$ candidate images $\{I_c^1, I_c^2, \dots, I_c^{k_1}\}$ via K-nearest
neighbor (KNN) search in the embedding space. To filter out incorrect retrievals, we perform geometric verification using a lightweight sparse matcher $\mathcal{M}_{\text{sparse}}$. 
For each image pair $\{I_q, I_c^i\}$, where $i\in\{1, \dots, k_1\}$, we computer the number of inlier correspondences under epipolar geometry constraints via a sparse feature matcher $\mathcal{M}_{\text{sparse}}$. The candidate image $I_{c}^c$ with the highest inlier number $N$ is selected as the initial retrieval result.

\para{Fine Viewpoint Retrieval.} 
If the inlier count $N$ falls below a threshold, indicating limited co-visibility, we leverage FGS’s NVS capability to find a novel view that is closer to the query image.
Specifically, we generate $k_2$ virtual views by randomly perturbing the pose of $I_{c}^c$ within a range of $a^\circ$ and $b$ m.
Gaussian map $\mathcal{G}$ renders virtual keyframes $\{I_v^1, I_v^2, \dots, I_v^{k_2}\}$  corresponding to these views.
We then re-run image retrieval to obtain $k_3$ new candidates frames and perform geometric verification again to select the best-aligned virtual image $I_{c}^f$, where $k_3 < k_1 \leq 10 < k_2 \leq 150$. 
Owing to the controlled number of geometric verification operations and the progressively shrinking search space, the proposed C2F retrieval pipeline remains \textit{highly efficient}.

\subsection{Hybrid Coarse-to-Fine Feature Matching}
\label{3.3}
Rendered images from FGS provide novel viewpoints that enable pose refinement, but they often contain artifacts.
Extracting features from these rendered images for feature matching often leads to unstable correspondences.
Thanks to the end-to-end feature rendering in FGS, the accumulation of errors during feature extraction is significantly reduced.
However, the rendered features from FGS and the query features from the image encoder $e$ exhibit a feature gap due to their different origins, which hinders fine pixel-level geometric relations.
To address this issue, we introduce a hybrid feature matcher $\mathcal{M}_{\text{hybrid}}$ that follows a C2F matching pipeline, which uses rendered features for coarse matching, and refines the coarse correspondences with a pretrained matcher in the fine matching stage.

\para{Coarse Feature Matching.} 
With a given estimated pose $\hat{p}_i$, we conduct coarse matching between the query feature map $F_t$ and the rendered feature map $F_r^{\text{high}}$ at the low resolution of $C \times H/8 \times W/8$.
During coarse matching, the coarse similarity matrix $S_c$ between the query and rendered feature maps is computed as follows:
\vspace{-6pt}
\begin{equation}
\label{sc}
S_c = \frac{1}{\tau} \cdot \left\langle F_t, F_r^{\text{high}} \right\rangle,
\vspace{-6pt}
\end{equation}
where $\tau$ is a temperature parameter, and $\langle \cdot, \cdot \rangle$ denotes the inner product between feature maps.
To derive matching probabilities, we apply a row-wise and column-wise softmax to obtain directional matching probability matrices:
\begin{equation}
P_{q \rightarrow r} = \text{Softmax}_{\text{row}}(S_c),
P_{r \rightarrow q} = \text{Softmax}_{\text{col}}(S_c).
\end{equation}
We then apply mutual nearest neighbor (MNN) filtering to $P_{q \rightarrow r}$ and $P_{r \rightarrow q}$ to establish coarse correspondences 
$\mathcal{C}_{q,r}^c$, which can be defined as:
\begin{align}
\mathcal{C}_{q,r}^c = \ & \left\{ (i, j) \,\middle|\, 
P^{q \rightarrow r}_{i,j} = \max_k P^{q \rightarrow r}_{i,k}, \ 
P^{q \rightarrow r}_{i,j} \geq \theta \right\} \nonumber \\
 \cap &\left\{ (i, j) \,\middle|\, 
P^{r \rightarrow q}_{i,j} = \max_k P^{r \rightarrow q}_{k,j}, \ 
P^{r \rightarrow q}_{i,j} \geq \theta \right\},
\end{align}
where $\theta$ is a confidence threshold that removes low-probability matches.

\para{Fine Feature Matching.} 
In the fine matching stage, coarse correspondences $\mathcal{C}_{q,r}^c$ are refined using fine features provided by a semi-dense matcher $\mathcal{M}_{\text{semi}}$. 
The semi-dense matcher $\mathcal{M}_{\text{semi}}$ is trained on a large-scale dataset annotated with accurate correspondences, enabling its fine feature to perform accurate geometric matching and generalize well across diverse scenes.
In our implementation, we use JamMa \cite{lu2025jamma} as the $\mathcal{M}_{\text{semi}}$, which demonstrates strong robustness in 2D–2D matching, while its Mamba-based architecture ensures both high efficiency and low memory usage. 

Given the rendered image $I_r$ and query image $I_q$, we first extract the fine feature via the fine feature encoder $E_{f}$ of the semi-dense matcher $\mathcal{M}_{\text{semi}}$:
\begin{equation}
    F_q^f, F_r^f = E_{f}(I_q, I_r), \quad F_q^f, F_r^f \in \mathbb{R}^{C^f \times H/2 \times W/2}.
\end{equation}
For each coarse correspondence, we crop a pair of $W \times W$ feature windows $\hat{F}_q^f, \hat{F}_r^f \in \mathbb{R}^{M \times W^2 \times C^f}$ from the fine features $F_q^f$ and $F_r^f$, where $M$ is the number of coarse matches. 
Following a similar strategy as Eq. \ref{sc}, we compute the correlation matrix $S_f$, and the fine matching probability matrix $P_f$ can be computed as:
\begin{equation}
P_f = \text{Softmax}_{\text{row}}(S_f) \cdot \text{Softmax}_{\text{col}}(S_f).
\end{equation}
The fine feature decoder $D_f$ of the semi-dense matcher $\mathcal{M}_{\text{semi}}$, integrated with a MNN filtering and a sub-pixel refinement module, is then used to compute the fine correspondences $\mathcal{C}_{q,r}^f$ as: $\mathcal{C}_{q,r}^f = D_{f}(P_f)$.

\begin{table*}
\footnotesize
\centering
\caption{\textbf{Relocalization results on 7-Scenes dataset.} We report the median translation (cm) and rotation (°) errors for each scene. The best and second-best results are \hl{\textbf{bolded}} and \underline{underlined}, respectively.}
\label{7-scenes result}
\begin{tabular}{@{}cl||cccccccc@{}}
\toprule

\multicolumn{1}{l}{} & \textbf{Methods} & \textbf{\textit{Chess}} & \textbf{\textit{Fire}} & \textbf{\textit{Heads}} & \textbf{\textit{Office}} & \textbf{\textit{Pumpkin}} & \textbf{\textit{RedKitchen}} & \textbf{\textit{Stairs}} & \textbf{Avg.↓} \\ \midrule
\multirow{4}{*}{\rotatebox{90}{Str.}}                             
& AS (SIFT)            & 3/0.87     & 2/1.01     & 1/0.82     & 4/1.15     & 7/1.69     & 5/1.72     & 4/1.01     & 3.71/1.18 \\ 
& HLoc (SP+SG) & 2.39/0.84 & 2.29/0.91  & 1.13/0.77  & 3.14/0.92  & 4.92/1.30  & 4.22/1.39  & 5.05/1.41 & 3.31/1.08 \\ 
& DVLAD+R2D2   & 2.56/0.88 & 2.21/0.86 & 0.98/0.75 & 3.48/1.00 & 4.79/1.28 & 4.21/1.44 & 4.60/1.27 & 3.26/1.07 \\
& \textbf{SplatHLoc$_\text{init}$ (Ours)} & 1.14/0.36 & 1.50/0.53 & 0.98/0.69 & 3.33/0.66 & 2.67/0.53 & 2.51/0.57 & 4.40/1.14 & 2.36/0.64 \\ \midrule

\multirow{4}{*}{\rotatebox{90}{Reg.}}           
& DSAC*               & 0.50/0.17 & 0.78/0.29 & 0.50/0.34 & 1.19/0.35 & 1.19/0.29 & 0.72/0.21 & 2.65/0.78 & 1.07/0.35 \\ 
& ACE                 & 0.55/0.18  & 0.83/0.33  & 0.53/0.33  & 1.05/0.29  & 1.06/0.22  & 0.77/0.21  & 2.89/0.81  & 1.10/0.34 \\ 
& NBE+SLD             & 0.6/0.18   & 0.7/0.26   & 0.6/0.35   & 1.3/0.33   & 1.5/0.33   & 0.8/0.19   & 2.6/0.72   & 1.16/0.34 \\
& NeuMap              & 2/0.81     & 3/1.11     & 2/1.17     & 3/0.98     & 4/1.11     & 4/1.33     & 4/1.12     & 3.14/0.95 \\ \midrule

\multirow{4}{*}{\rotatebox{90}{\shortstack{Ren.\\(NeRF)}}} 
& DFNetS$_{50}$  & 2/0.57     & 2/0.74     & 2/1.28     & 2/0.56     & 2/0.55     & 2/0.57     & 5/1.28     & 2.43/0.79 \\
& CrossFire           & 1/0.4      & 5/1.9      & 3/2.3      & 5/1.6      & 3/0.8      & 2/0.8      & 12/1.9     & 4.43/1.38 \\
& PNeRFLoc            & 2/0.80     & 2/0.88     & 1/0.83     & 3/1.05     & 6/1.51     & 5/1.54     & 32/5.73    & 7.29/1.76 \\
& NeRFMatch           & 0.95/0.30  & 1.11/0.41  & 1.34/0.92  & 3.09/0.87  & 2.21/0.60  & 1.03/0.28  & 9.26/1.74  & 2.71/0.73 \\ \midrule

\multirow{6}{*}{\rotatebox{90}{\shortstack{Ren.\\(GS)}}} 
& GSplatLoc (Fine)    & 0.39/0.13   & 0.91/0.29   & 0.94/0.50  & 1.41/0.32  & 1.41/0.26  & 1.32/0.29  & 3.44/0.82  & 1.40/0.37 \\
& ACE+GS-CPR          & 0.5/0.15   & 0.6/0.25   & \underline{0.4}/0.28  & 0.9/0.26  & 1.0/0.23  & 0.7/0.17  & 1.4/0.42  & 0.78/0.25 \\
& LoGS              & 0.4/\underline{0.10}  & 0.6/\hl{\textbf{0.18}}  & 0.5/\underline{0.26}  & 0.7/0.22  & \hl{\textbf{0.7}}/0.22  & \underline{0.5/0.14}  & 1.6/0.43  & 0.76/0.24 \\
& RAP$_\text{ref}$    & \underline{0.33}/0.11   & \underline{0.51}/0.21   & \underline{0.40}/0.29  & \hl{\textbf{0.59/0.17}}  & \underline{0.83/0.20}  & \underline{0.50}/\hl{\textbf{0.12}}  & \underline{1.11/0.32}  & \underline{0.61/0.20} \\
& STDLoc              & 0.46/0.15  & 0.57/0.24  & 0.45/\underline{0.26}  & 0.86/0.24  & 0.93/0.21  & 0.63/0.19  & 1.42/0.41  & 0.76/0.24 \\
& \textbf{SplatHLoc (Ours)} &
  \hl{\textbf{0.28/0.09}} &
  \hl{\textbf{0.44}}/\underline{0.19} &
  \hl{\textbf{0.29/0.18}} &
  \underline{0.62/0.18} &
  \hl{\textbf{0.70/0.15}} &
  \hl{\textbf{0.49/0.12}} &
  \hl{\textbf{1.03/0.30}} &
  \hl{\textbf{0.55/0.17}} \\ 
  \bottomrule
\end{tabular}
\end{table*}

\begin{figure*}[htbp]
    \centering
    \includegraphics[width=0.98\textwidth]{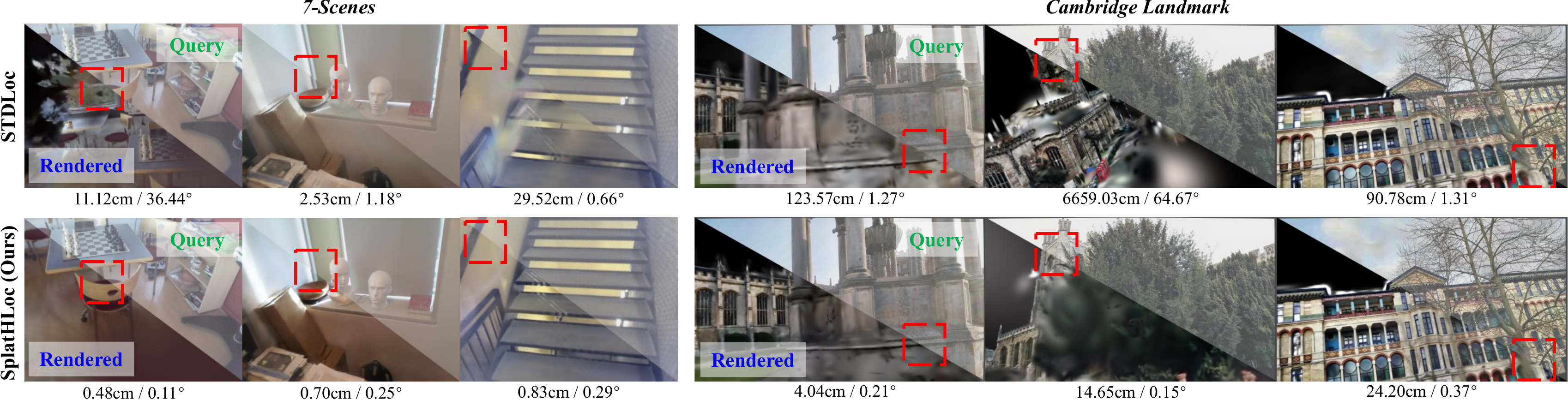}
    \caption{\textbf{Visualization of the relocalization errors.} Each subfigure is divided by a diagonal: the top-right part shows the \textcolor{newgreen}{query image} in grayscale, while the bottom-left part shows the \textcolor{blue}{rendered image} from the estimated pose. The \textcolor{red}{red} dashed boxes highlight regions with pronounced visual differences in each column. More visualizations and details are in the supplementary material.}
    \vspace{-11pt}
    \label{fig: viz_loc}
\end{figure*}

\subsection{Relocalization Process}
\label{3.4}
Our relocalization approach follows a hierarchical framework.
First, we retrieve a reference image that is similar to the query image as described in Section \ref{3.2}, along with its associated camera intrinsics $K \in \mathbb{R}^{3 \times 3}$ and extrinsics $T \in \mathbb{R}^{4 \times 4}$.
Using these camera parameters, we render the color image $I_r$ and depth map $D_r$.
We then use our proposed hybrid feature matcher $\mathcal{M}_{\text{hybrid}}$ proposed in Section \ref{3.3} to establish 2D-2D correspondences $\mathcal{C}_{q,r}^f$ between the query image $I_q$ and the rendered image $I_r$, where coarse matching is performed between the query feature map $F_t$ and the rendered feature map $F_r^{\text{high}}$: 
\begin{equation}
\label{match}
    \mathcal{C}_{q,r}^f = \mathcal{M}_{\text{hybrid}}(F_t, F_r^{\text{high}}, I_q, I_r).
\end{equation}
Each 2D–2D correspondence $(\mathbf{x}_q, \mathbf{x}_r) \in \mathcal{C}_{q,r}^f$ is lifted to a 2D–3D correspondence $(\mathbf{x}_q, \mathbf{X}_r)$ via:
\vspace{-3pt}
\begin{equation}
\label{lift2d3d}
\mathbf{X}_r = T \cdot \left( D_r(\mathbf{x}_r) \cdot K^{-1} \begin{bmatrix} \mathbf{x}_r \\ 1 \end{bmatrix} \right).
\vspace{-3pt}
\end{equation}
Finally, we feed the 2D-3D correspondences into the PnP \cite{gao2003complete} algorithm with RANSAC \cite{fischler1981random} to estimate the 6-DOF \textit{initial} pose $\hat{p_i} \in \mathbb{R}^{4 \times 4}$ of the query image.

To further refine the estimated pose, we render new color and depth maps from the estimated pose, perform image matching with the query image following Eq. (\ref{match}) and Eq. (\ref{lift2d3d}), and re-estimate a refined pose $\hat{p_r}^n\in \mathbb{R}^{4 \times 4}$, where $n$ denotes the refinement iteration number. If $n = 1$, the initial pose $p_i$ is used as input. Otherwise, we adopt the pose $\hat{p}_r^{n-1}$ from the $(n-1)$-th refinement.

\vspace{-3pt}
\section{Experiments}

\vspace{-3pt}
Here, we evaluate the effectiveness of the proposed visual relocalization method.
We begin by presenting the experimental setup in Section \ref{4.1}.
Following this, we provide a detailed comparative analysis of relocalization accuracy in Section \ref{4.2} and complexity in Section \ref{4.3} against previous SotA methods.
Finally, we conduct ablation studies to investigate the impact of our contributions in Section \ref{4.4}.

\vspace{-3pt}
\subsection{Experimental Setup}
\label{4.1}

\vspace{-3pt}
\para{Datasets.}
We evaluate the performance of SplatHLoc on three well-known public visual relocalization datasets spanning indoor and outdoor environments. 
The \textit{7-Scenes} \cite{shotton2013scene} dataset consists of seven indoor scenes captured using a handheld Kinect RGB-D camera at a resolution of $640 \times 480$, representing small-scale indoor environments with diverse texture patterns and varying geometric structures.
The \textit{12-Scenes} \cite{valentin2016learning} dataset provides RGB-D scans of 12 rooms across 4 large indoor scenes, covering a variety of lighting conditions, layouts, and object arrangements.
The \textit{Cambridge Landmarks} \cite{kendall2015posenet} dataset comprises five outdoor scenes captured using mobile phones. It presents several real-world challenges, including large-scale environments, dynamic objects, illumination changes, and motion blur.

\para{Metric.}
We report two evaluation metrics widely used in prior research \cite{huang2025sparse, liu2025gscpr, pietrantoni2025gaussian, botashev2024gsloc, sidorov2025gsplatloc, cheng2025logs} for comparing the performance of different methods: (1) the median translation and rotation error, and (2) the recall rate, defined as the percentage of query images with translation and rotation errors falling below predetermined thresholds.

\para{Implementation Details.}
In the \textit{training stage}, we use SuperPoint \cite{detone2018superpoint} as the image encoder $e$, yielding a feature dimension of $C = 256$, while $C'$ is set to 64.
Our training parameters follow those of the vanilla 3DGS \cite{kerbl20233d}. 
Each scene is trained for 30,000 steps with a learning rate of 0.001 for the feature field training, and $\gamma = 1$ in Eq. \ref{loss}, $\lambda = 0.2$ in Eq. \ref{eq:Lrgb}.
We use gsplat \cite{ye2025gsplat} for the optimization.
In the \textit{relocalization stage}, we use MixVPR \cite{ali2023mixvpr} as the VPR model $\mathcal{V}$ for efficient image retrieval, and employ SuperPoint \cite{detone2018superpoint} with LightGlue \cite{lindenberger2023lightglue} as the sparse matcher $\mathcal{M}_{\text{sparse}}$ for geometric verification.
Pose estimation is performed using PoseLib \cite{PoseLib}.
The iteration number $n$ for pose refinement is set to 4 for 7-Scenes and 12-Scenes, and set to 2 for Cambridge Landmarks, in order to ensure convergence of the estimated pose. 
More parameters of the proposed method are shown in the supplementary material.
All experiments are conducted on a server with a single NVIDIA RTX 3090 GPU.

\begin{figure*}[htbp]
    \centering
    \includegraphics[width=0.98\textwidth]{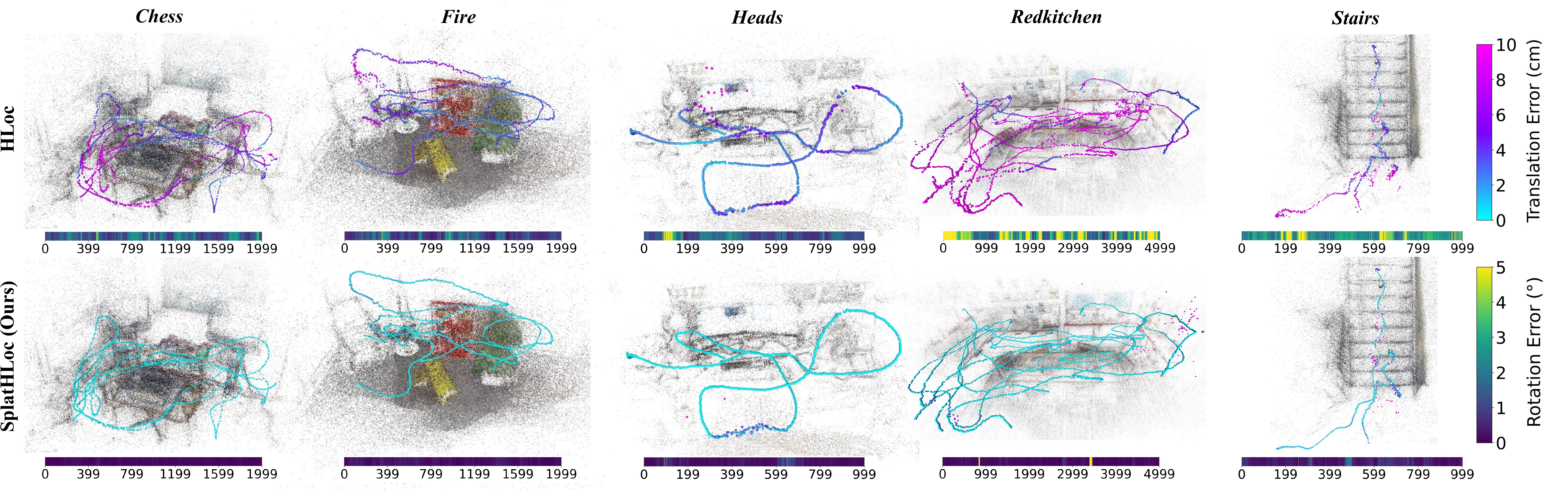}
    \caption{\textbf{Qualitative comparison of camera pose estimation errors} between HLoc \cite{sarlin2019coarse} and our proposed SplatHLoc across five scenes from the 7-Scenes dataset. Visualizations of the remaining two scenes and more details are provided in the supplementary material. For each scene, we visualize the reconstructed point cloud map together with the trajectory of query images. Trajectory colors denote position error, while the color bar below shows rotation errors, with numbers indicating image indices.}
    \vspace{-11pt}
    \label{fig: traj_loc}
\end{figure*}

\subsection{Relocalization Accuracy}
\label{4.2}
To assess the improved relocalization accuracy of our approach, we conduct extensive experiments to compare it against a broad set of visual relocalization methods.
These methods include \textit{structure-based methods}: AS (SIFT) \cite{sattler2012improving, lowe2004distinctive}, HLoc (SP+SG) \cite{sarlin2019coarse, detone2018superpoint, sarlin2020superglue}, DVLAD+R2D2 \cite{torii201524, revaud2019r2d2}; \textit{regression-based methods}: DSAC* \cite{brachmann2021visual}, ACE \cite{brachmann2023accelerated}, NBE+SLD \cite{do2022learning}, GLACE \cite{wang2024glace}, NeuMap \cite{tang2023neumap}, Marepo \cite{chen2024map}; \textit{render-based methods with NeRF}: DFNetS$_{50}$ \cite{chen2024neural}, CrossFire \cite{moreau2023crossfire}, PNeRFLoc \cite{zhao2024pnerfloc}, NeRFMatch \cite{zhou2024nerfect}; and \textit{render-based methods with GS}: GSplatLoc \cite{sidorov2025gsplatloc}, ACE+GS-CPR \cite{brachmann2023accelerated, liu2025gscpr}, LoGS \cite{cheng2025logs},  RAP$_\text{ref}$ \cite{li2025adversarial}, STDLoc \cite{huang2025sparse}.

\para{Indoor Relocalization.} We report the median translation and rotation errors on 7-Scenes dataset in Table \ref{7-scenes result}.  
Structure-based methods are a common choice for initial pose estimation, but they tend to perform worse than regression-based alternatives in indoor environments.
Notably, SplatHLoc$_\text{init}$, our proposed initial pose estimation method, outperforms other structure-based methods and even surpasses NeuMap, a representative regression-based method.
Building upon SplatHLoc$_\text{init}$ with an additional pose refinement step, SplatHLoc significantly improves relocalization accuracy and achieves the best performance among all evaluated methods.

\begin{table}[t]
\renewcommand{\arraystretch}{1.15} 
    \centering
    \setlength{\tabcolsep}{3.5pt}
    \caption{\textbf{Relocalization results on the 12-Scenes dataset.} We report the median translation (cm) and rotation errors (°), and the percentage of query images below 2 cm, 2° pose error.}
    \label{12Scenes result}
    \scalebox{0.8}{
    \begin{tabular}{l||cc}
    \toprule
    \textbf{Methods}  & \textbf{Avg. Err [cm/°]} $\downarrow$  & \textbf{$R$@[2cm, 2°]} $\uparrow$  \\
    \midrule
    DSAC* (\textit{T-PAMI'21})      & \underline{0.5}/0.25  & 96.7  \\
    Marepo (\textit{CVPR'24})      & 2.1/1.04  & 50.4  \\
    Marepo+GS-CPR (\textit{ICLR'25})      & 0.7/0.28 & 90.9  \\
    ACE (\textit{CVPR'23})     & 0.7/0.26  & 97.2  \\
    ACE+GS-CPR (\textit{ICLR'25})      & \underline{0.5/0.21}  & \hl{\textbf{98.7}}  \\
     \textbf{SplatHLoc$_\text{init}$ (Ours)}   & 1.34/0.37 & 71.3   \\
     \textbf{SplatHLoc (Ours)}  & \hl{\textbf{0.3/0.14}} & \underline{97.3}  \\
    \bottomrule
    \end{tabular}}
    \vspace{-11pt}
\end{table}

We also conduct a quantitative evaluation on the 12-Scenes dataset. As shown in Table \ref{12Scenes result}, our SplatHLoc outperforms other methods in terms of median translation and rotation errors. Even though the initial percentage of query images below 2 cm, 2° pose error is 19.6\% lower than ACE, SplatHLoc’s $R$@[2cm, 2°] is still very close to ACE+GS-CPR (only 1.4\% lower) and higher than Marepo+GS-CPR.

\para{Outdoor Relocalization.} 
We report the median translation and rotation errors on the Cambridge Landmarks dataset in Table \ref{cam_result}. 
In large-scale outdoor environments, SplatHLoc$_\text{init}$ achieves nearly half the error compared to render-based methods with NeRF, such as DFNetS$_{50}$, CrossFire, and PNeRFLoc.
Table \ref{cam_result} also shows that SplatHLoc remains competitive with the current SotA method, STDLoc. 
Compared to its initial pose estimates from SplatHLoc$_\text{init}$, SplatHLoc consistently achieves significant improvement across large-scale scenes.
This demonstrates the effectiveness of SplatHLoc in challenging outdoor environments.

\para{Visualization.} In Fig. \ref{fig: viz_loc}, we show some qualitative examples for both indoor and outdoor relocalization.
The visualization in Fig. \ref{fig: traj_loc} shows that our method, SplatHLoc, which enables pose refinement, achieves more robust and accurate relocalization performance than HLoc, despite both being hierarchical frameworks.

\begin{table}
\scriptsize
\setlength{\tabcolsep}{1pt}
\centering
\caption{\textbf{Relocalization results on Cambridge Landmarks dataset.} We report the median translation (cm) and rotation (°) errors of different methods.}
\label{cam_result}
\begin{tabular}{@{\hspace{5pt}}cl||cccccc@{\hspace{4pt}}}
\toprule
\multicolumn{1}{l}{} & \textbf{Methods} & \textbf{\textit{Court}} & \textbf{\textit{College}} & \textbf{\textit{Hospital}} & \textbf{\textit{Shop}} & \textbf{\textit{Church}} & \textbf{Avg.↓} \\ \midrule

\multirow{3}{*}{\rotatebox{90}{Str.}}        
& AS (SIFT)     & 24/0.13 & 13/0.22 & 20/0.36 & \underline{4}/0.21 & 8/0.25 & 14/0.23 \\
& HLoc (SP+SG)  & 18/0.11 & \hl{\textbf{11}}/0.20 & 15/0.31 & \underline{4}/0.20 & 7/0.22 & 11/0.21 \\ 
& \textbf{SplatHLoc$_\text{init}$ (Ours)} & 31/0.15
& 19/0.26 
& 24/0.44
& 6/0.19 
& 6/0.21
& 17/0.25 \\ \midrule

\multirow{4}{*}{\rotatebox{90}{Reg.}}           
& DSAC*         & 33/0.21 & 18/0.31 & 21/0.40 & 5/0.24 & 15/0.51 & 19/0.33 \\
& ACE           & 43/0.2 & 28/0.4 & 31/0.6 & 5/0.3 & 18/0.6 & 25/0.4 \\
& GLACE         & 19/0.12 & 19/0.32 & 18/0.42 & \underline{4}/0.23 & 8/0.29  & 14/0.28 \\
& NeuMap        & \hl{\textbf{6}}/0.10 & 14/\underline{0.19} & 19/0.36 & 6/0.25 & 17/0.53 & 12/0.29 \\ \midrule

\multirow{4}{*}{\rotatebox{90}{\shortstack{Ren.\\(NeRF)}}} 
& DFNetS$_{50}$ & - & 37/0.54 & 52/0.88 & 15/0.53 & 37/1.14 & 35/0.77 \\
& CrossFire     & - & 47/0.7 & 43/0.7 & 20/1.2 & 39/1.4 & 37/1.00 \\
& PNeRFLoc      & 81/0.25 & 24/0.29 & 28/0.37 & 6/0.27 & 40/0.55 & 36/0.35 \\
& NeRFMatch     & 20/0.09 & \underline{13}/0.23 & 21/0.38 & 8/0.40 & 11/0.35 & 15/0.29 \\ \midrule

\multirow{6}{*}{\rotatebox{90}{\shortstack{Ren.\\(GS)}}} 
& GSplatLoc (Fine)  & - & 31/0.49 & 16/0.68 & \underline{4}/0.34 & 14/0.42 & 16/0.48 \\
& ACE+GS-CPR   & - & 20/0.29 & 21/0.40 & 5/0.24 & 13/0.40 & 15/0.33 \\
& LoGS  & 13/0.09 & \hl{\textbf{11}}/\underline{0.19} & 15/0.31 & \underline{4}/0.19 & 7/0.20 & \underline{10}/0.20 \\
& RAP$_\text{ref}$  & 23/0.15 & 15/0.25 & 18/0.38 & 5/0.25 & 9/0.23 & 14/0.25 \\
& STDLoc  & 16/\hl{\textbf{0.06}} & 15/\hl{\textbf{0.17}} & \underline{12/0.21} & \hl{\textbf{3}}/\underline{0.13} & \underline{5/0.14} & \underline{10/0.14} \\
& \textbf{SplatHLoc (Ours)} & \underline{14}/\underline{0.07} 
& 17/0.21 
& \hl{\textbf{8/0.15}} 
& \hl{\textbf{3/0.12}} 
& \hl{\textbf{4/0.12}} 
& \hl{\textbf{9/0.13}} \\
\bottomrule
\end{tabular}
\vspace{-11pt}
\end{table}

\begin{figure}[htbp]
    \centering
    \includegraphics[width=0.47\textwidth]{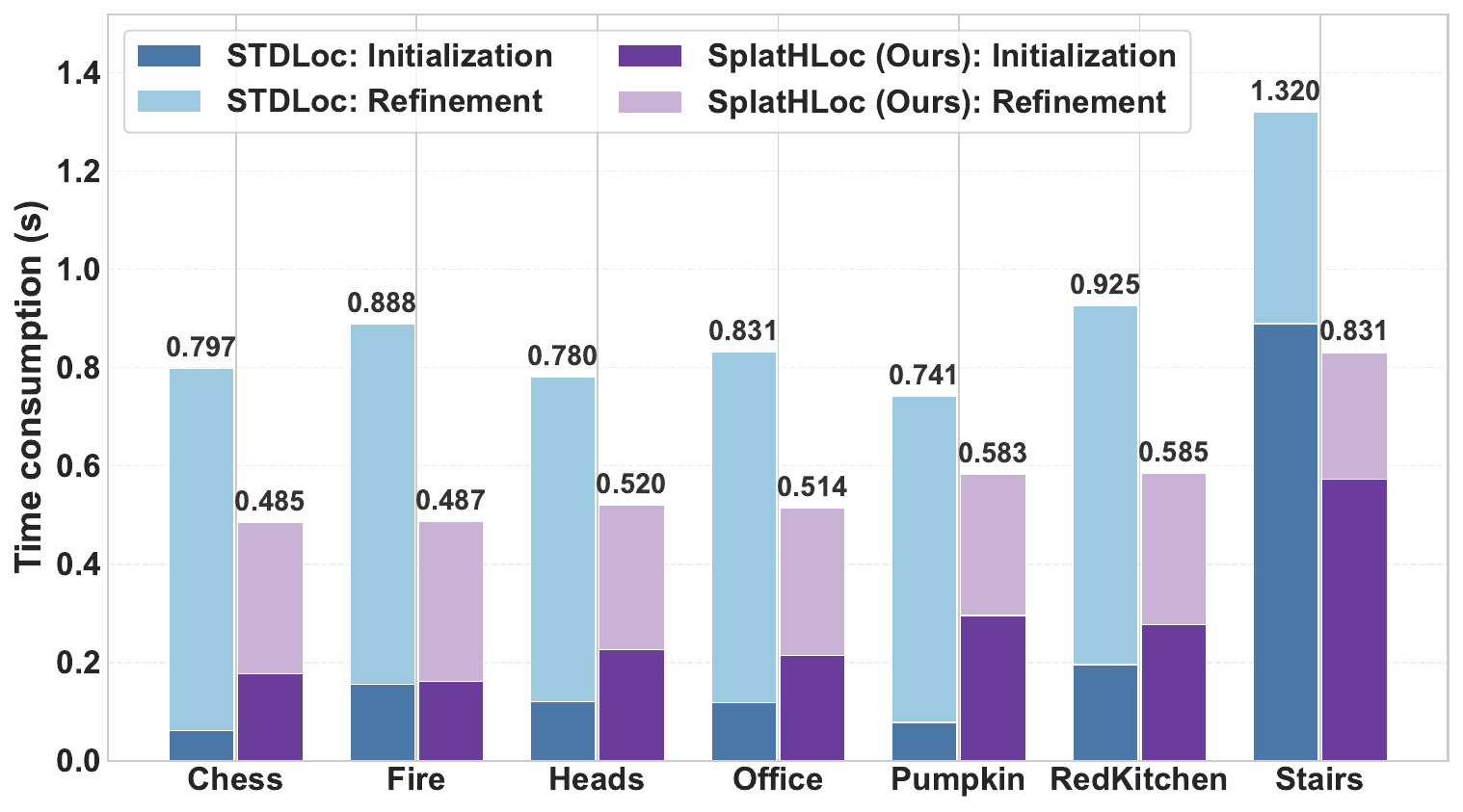}
    \caption{\textbf{Runtime analysis.} We report the average relocalization time per query on the 7-Scenes dataset for STDLoc versus our proposed SplatHLoc. The runtime is divided into two parts: initial pose estimation and iterative pose refinement. Both methods perform four rounds of pose refinement.}
    \vspace{-6pt}
    \label{fig: runtime}
\end{figure}

\subsection{Complexity Analysis}
\label{4.3}
We conduct experiments to evaluate the complexity of SplatHLoc and compare it against a SotA method STDLoc. 
As shown in Fig. \ref{fig: runtime}, SplatHLoc consistently achieves lower runtime than STDLoc across all seven scenes.
\textit{For initial pose estimation time}, SplatHLoc is slightly higher than STDLoc in the first six scenes.
However, STDLoc requires sampling Gaussian spheres and training a scene-specific detector, introducing overhead that hampers scalability. 
In the low-texture \textit{Stairs} scene, its direct 2D–3D matching for initialization becomes slow, exceeding even the total runtime of SplatHLoc.
\textit{For iterative pose refinement time}, SplatHLoc is nearly twice as fast as STDLoc. This speedup stems from two factors: (1) SplatHLoc only renders low-dimensional features and then lifts them with a feature decoder $d$, whereas STDLoc directly renders high-dimensional feature maps, and (2) during fine matching, SplatHLoc operates on feature maps at 2× downsampled image resolution, whereas STDLoc runs at full resolution.

We also compare STDLoc and our SplatHLoc in the \textit{FGS training stage} as shown in Table \ref{map}. Benefiting from the feature decoder $d$, SplatHLoc uses lower per-Gaussian feature dimensionality than STDLoc (64 vs. 256), leading to significant reductions in FGS map size, mapping time, and peak GPU memory. 
Moreover, SplatHLoc’s feature decoder is a single-layer CNN trained jointly with the Gaussian map, which has almost no impact on the mapping time.

\subsection{Ablation Study}
\label{4.4}

In Table \ref{tab:ablation}, we conduct ablation studies on the \textit{Stairs} scene, a typical low-texture scene from the 7-Scenes dataset,  to investigate the impact of all the proposed components in our SplatHLoc.
Setup \hyperref[case: I]{I} is our baseline, which use MixVPR for image retrieve, and use SuperPoint (SP) with LightGlue (LG) as a feature matcher $\mathcal{M}$ for initial pose estimate and iterative pose refinement.
In Setup \hyperref[case: II]{II}, we incorporate the adaptive strategy introduced in Section \ref{3.2} into the image retrieval process.
In the following setups, we replace the feature matcher with the hybrid feature matcher introduced in Section \ref{3.3}.
Comparing Setup \hyperref[case: I]{I} vs. \hyperref[case: II]{II}, and Setup \hyperref[case: III]{III} vs. \hyperref[case: IV]{IV}, the latter adds adaptive retrieval on top of the former.  
The experiments show that, under different feature matchers, adaptive retrieval can consistently boost the relocalization performance.
Comparing Setup \hyperref[case: I]{I} vs. \hyperref[case: III]{III} and Setup \hyperref[case: II]{II} vs. \hyperref[case: IV]{IV},
the results show that the hybrid feature matcher consistently outperforms SP+LG. 

\begin{table}[t]
    \small
    \renewcommand{\arraystretch}{1.15}
    \centering
    \setlength{\tabcolsep}{3.5pt}
    \caption{\textbf{Mapping analysis.} We report the FGS map size, mapping time and peak GPU memory on the \textit{Chess} scene.}
    \label{map}
    \scalebox{0.78}{
    \begin{tabular}{lccc}
    \toprule
    \textbf{Methods} & \textbf{Map Size (MB)} &\textbf{Mapping Time (Min)} & \textbf{GPU Mem. (GB)} \\
    \midrule
    STDLoc & 904.02 & $\sim$146 & $\sim$12 \\
    \textbf{SplatHLoc (Ours)} & 353.21 & $\sim$46 & $\sim$4  \\
    \bottomrule
    \end{tabular}}
\end{table}

\begin{table}[t]
    \small
    \centering
    \caption{\textbf{Ablation study I.} We investigate the impact of the proposed Adaptive C2F Retrieval and Hybrid C2F Matching.}
    \label{tab:ablation}
    \scalebox{0.8}{
    {\renewcommand{\arraystretch}{1.2}
    \begin{tabular}{lccc}
    \toprule
    \textbf{Setups}\quad\quad\quad  on \textit{Stairs}  & \textbf{Avg. Err [cm/°]} $\downarrow$ & \textbf{$R$@[5cm, 5°]} $\uparrow$\\
    \midrule
    \rowcolor{gray!25}\label{case: I}I (Baseline): MixVPR + $\mathcal{M}$: SP+LG  & 1.82/0.49  & 75.5 \\
    \label{case: II}II: I + Adaptive Retrieval  &	1.57/0.45  & 80.5 \\
    \label{case: III}III: I + $\mathcal{M}$: Hybrid Matcher  & 1.14/0.33  & 84.0 \\
    \rowcolor{green!25} \label{case: IV}\textbf{IV (Ours):}  II + $\mathcal{M}$: Hybrid Matcher & \textbf{1.03/0.30} & \textbf{91.9} \\
    \bottomrule
    \end{tabular}}}
    \vspace{-11pt}
\end{table}

\begin{table}[ht]
    \centering
    \small
    \renewcommand{\arraystretch}{1.0}
    \setlength{\tabcolsep}{10pt}
    \caption{\textbf{Ablation study II.} We ablate the hybrid matcher by isolating rendered features’ coarse and fine contributions. FGS-Matcher is adopted from STDLoc, which uses rendered features in both coarse and fine matching stages.}
    \label{tab:c2f}
    \scalebox{0.7}{
    \begin{tabular}{l c c c}
    \toprule
    \multicolumn{1}{l}{\textbf{Datasets}} & \multicolumn{2}{c}{\textbf{7-Scenes}} & \multicolumn{1}{c}{\textbf{Cambridge}}\\
    \cmidrule(lr){1-1} \cmidrule(lr){2-3} \cmidrule(lr){4-4}
    \textbf{Methods} & \textbf{$R$@[2cm, 2$^\circ$]} $\uparrow$ & \textbf{Avg. Err [cm/°]} $\downarrow$ & \textbf{$R$@[5m, 10$^\circ$]} $\uparrow$\\
    \midrule
    $\mathcal{M}$: FGS-Matcher & 91.46 & 0.65/0.22 & 98.28\\
    \midrule
    $\mathcal{E}$: ELoFTR & 90.30 & 0.57/0.19 & 97.86 \\
    $\mathcal{M} + \mathcal{E_\text{coarse}}$ & 89.98$_{\,(-0.32)}$ & 0.64/0.21 & 96.72$_{\,(-1.14)}$ \\
    $\mathcal{M} + \mathcal{E_\text{fine}}$ & \textbf{93.55$_{\,(+3.25)}$} & \textbf{0.56/0.18} & \textbf{98.73$_{\,(+0.87)}$} \\
    \midrule
    $\mathcal{J}$: JamMa & 90.93 & 0.60/0.19 & 97.78\\
    $\mathcal{M} + \mathcal{J_\text{coarse}}$ & 90.32$_{\,(-0.61)}$ & 0.64/0.22 & 97.84$_{\,(+0.06)}$\\
    $\mathcal{M} + \mathcal{J_\text{fine}}$ & \textbf{93.84$_{\,(+2.91)}$} & \textbf{0.55/0.17} & \textbf{98.41$_{\,(+0.63)}$} \\
    \bottomrule
    \end{tabular}}
    \vspace{-11pt}
\end{table}

To further validate the effectiveness of the proposed hybrid matcher, we replace coarse and fine stages from FGS-Matcher \cite{huang2025sparse} with the corresponding modules in a semi-dense matcher, as shown in Table. \ref{tab:c2f}.
We employ two SotA semi-dense matchers, ELoFTR \cite{wang2024efficient} and JamMa \cite{lu2025jamma}. 
Using the coarse features from a semi-dense matcher for coarse matching and rendered features for fine matching underperforms both the standalone FGS-Matcher and the standalone semi-dense matcher. 
In contrast, the reversed pairing outperforms other methods.
These results confirm that rendered features mainly benefit the coarse matching stage, while fine matching is more accurate with features extracted from color images with a semi-dense matcher.

\section{Conclusion}

\para{Summary.} 
In this paper, we have presented a novel hierarchical visual relocalization method, SplatHLoc, which leverages Feature Gaussian Splatting as the scene representation.
Using synthesized color maps, SplatHLoc enables adaptive viewpoint retrieval, making the retrieved image more consistent in viewpoint with the query. 
Moreover, we observe that rendered features mainly take effect in the coarse matching stage.
By combining the synthesized feature maps for coarse matching with fine features from a semi-dense matcher for fine matching, the hybrid feature matching produces more accurate 2D–2D matches. 
Finally, the synthesized depth maps lift 2D–2D matches to 2D–3D for pose estimation.
Experimental results show that SplatHLoc outperforms SotA methods in both accuracy and efficiency, producing robust pose estimates under weak textures and illumination changes.
We believe that this study can inspire further research on visual relocalization.

\para{Limitation and Future Work.} 
The performance of our method depends on the Gaussian map quality, which is influenced by the number of images used for FGS mapping.
Future work includes: (1) Exploring the use of recent 3D reconstruction foundation models \cite{deng2025sail} in place of COLMAP to initialize Gaussian primitives. (2) Extending our method to large scale maps \cite{jiao2026opennavmap} by dividing the scene into blocks.

\section*{Acknowledgment} This work was supported in part by Shenzhen Science and Technology Program (No. SGDX20240115111759002), in part by Meituan Academy of Robotics Shenzhen, in part by the Shenzhen Association for Science and Technology (No. XHXS2025-003), and in part by High level of special funds (G03034K003) from Southern University of Science and Technology, Shenzhen, China.

{
    \small
    \bibliographystyle{ieeenat_fullname}
    \bibliography{main}
}

\clearpage

\setcounter{page}{1}
\maketitlesupplementary

\renewcommand{\thetable}{\Roman{table}}
\renewcommand{\thefigure}{\Roman{figure}}
\renewcommand{\theequation}{\Roman{equation}}
\renewcommand\thesection{\Alph {section}}
\setcounter{section}{0}
\setcounter{figure}{0}
\setcounter{table}{0}
\setcounter{equation}{0}


\appendix

\section*{Contents}

\vspace{0.5em}

\begin{itemize}[leftmargin=1.8em]
  \item[\textbf{A}] \hyperref[app:adv_view]{Details of Adaptive Viewpoint Retrieval}
  \item[\textbf{B}] \hyperref[app:scene_meta]{Scene Metadata}
  \item[\textbf{C}] \hyperref[app:mapping_quality]{Mapping Quality}
  \item[\textbf{D}] \hyperref[app:extra_exp]{Additional Experimental Results}
  \item[\textbf{E}] \hyperref[app:extra_vis]{Additional Visualization}
  \item[\textbf{F}] \hyperref[app:fgs_raster]{Gaussian Rasterization in FGS}
\end{itemize}

\section{Details of Adaptive Viewpoint Retrieval}\label{app:adv_view}

The algorithm of the adaptive viewpoint retrieval strategy is illustrated in Algorithm~\ref{alg:adaptive_c2f_retrieval}.

\para{Hyperparameters.}
The hyperparameters used in this retrieval process for indoor and outdoor scenes are summarized in Table~\ref{tab:k_values}.
These include the number of retrieved images $k_1$ for geometric verification in the coarse stage, the number of virtual views $k_2$ generated by pose perturbation, the number of candidate images $k_3$ verified in the fine stage, the perturbation ranges $(a^\circ, b\text{ m})$ for rotation and translation, and the inlier threshold $\mathcal{I}$ used during geometric verification.
Moreover, the images retrieved in the coarse stage exhibit strong viewpoint redundancy.
To avoid repeated verification on nearly identical views, we perform geometric verification only on every 10th retrieved image.

\begin{table}[H]
    \centering
    \caption{Hyperparameters in the adaptive retrieval process.}
    \label{tab:k_values}
    \resizebox{0.9\linewidth}{!}{
    \begin{tabular}{l|cccccc}
    \toprule
    Scene Type & $k_1$ & $k_2$ & $k_3$ & $a$ & $b$ & $\mathcal{I}$\\
    \midrule
    Indoor  & 10 & 150 & 5 & 5 & 0.5 & 150\\
    Outdoor & 10 & 100 & 5 & 5 & 0.8 & 300\\
    \bottomrule
    \end{tabular}}
\end{table}

\para{Perturbation sampling strategies}
We compare multiple perturbation strategies generated using different sampling distributions on the stairs scene. As shown in Table \ref{tab:distribution}, Normal and Random perturbations consistently outperform Uniform sampling in both accuracy and robustness. Normal sampling yields the best localization success rate, whereas Random provides a favorable trade-off between accuracy and runtime.

\begin{table}[t]
    \small
    \centering
    \caption{Comparison of Uniform, Normal, and Random perturbation strategies. Time denotes the average runtime required for the initial localization.}
    \label{tab:distribution}
    \resizebox{0.99\linewidth}{!}{
    {\renewcommand{\arraystretch}{1.2}
    \begin{tabular}{lccc}
    \toprule
    \textbf{Distributions}  & \textbf{Avg. Err [cm/°]} $\downarrow$ & \textbf{$R$@[5cm, 5°]} $\uparrow$ & \textbf{Time (s)} $\downarrow$\\
    \midrule
    Uniform & 1.11/0.33  & 85.3 &  0.60\\
    Normal & \textbf{1.03/0.30}  & \textbf{92.8} & 0.60\\
    Random & \textbf{1.03/0.30} & 91.9  & \textbf{0.57}\\
    \bottomrule
    \end{tabular}}}
\end{table}

\begin{figure}[htbp]
    \centering
    \includegraphics[width=0.47\textwidth]{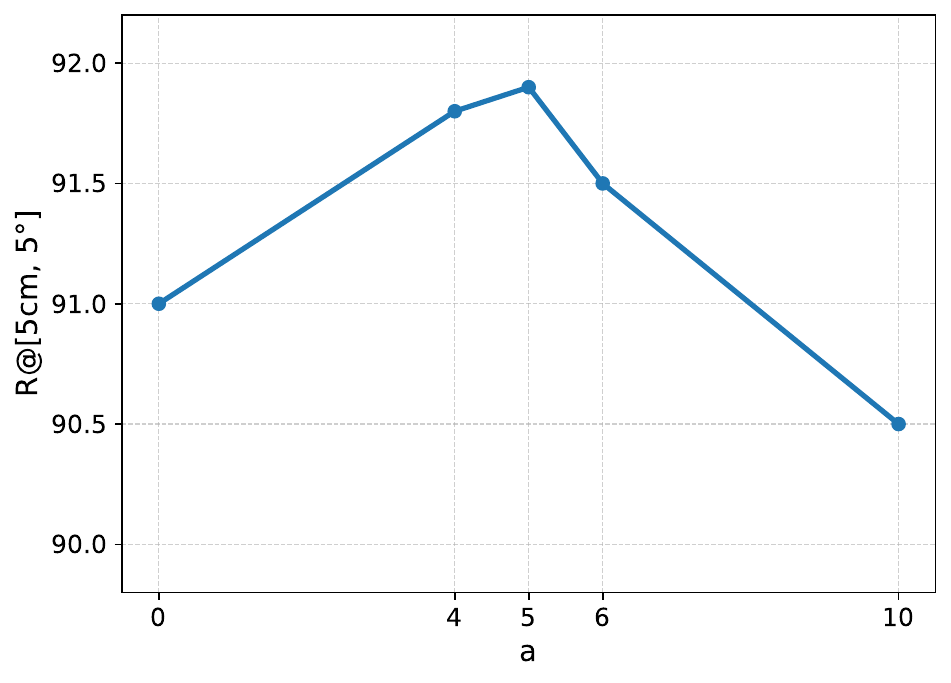}
    \caption{The R@[5cm, 5°] accuracy under different perturbation angle values $a$, with the perturbation distance $b$ fixed to $0.5 \text{ m}$.}
    \label{fig:a_stairs}
\end{figure}

\begin{figure}[htbp]
    \centering
    \includegraphics[width=0.47\textwidth]{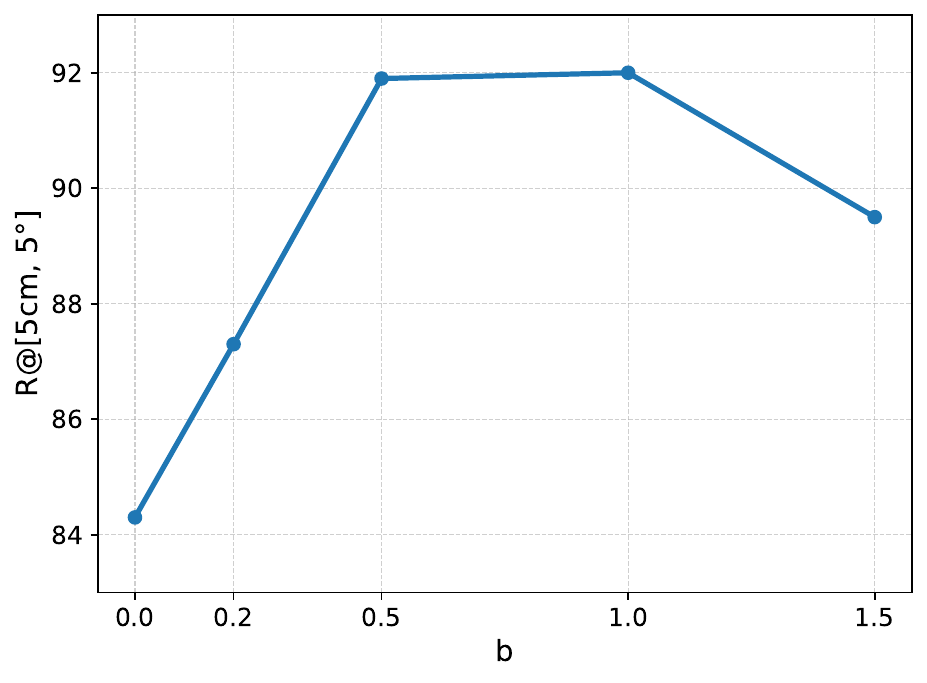}
    \caption{The R@[5cm, 5°] accuracy under different perturbation distance values $b$, with the perturbation angle $a$ fixed to 5°.}
    \label{fig:b_stairs}
    \vspace{-11pt}
\end{figure}

\begin{figure}[htbp]
    \centering
    \includegraphics[width=0.47\textwidth]{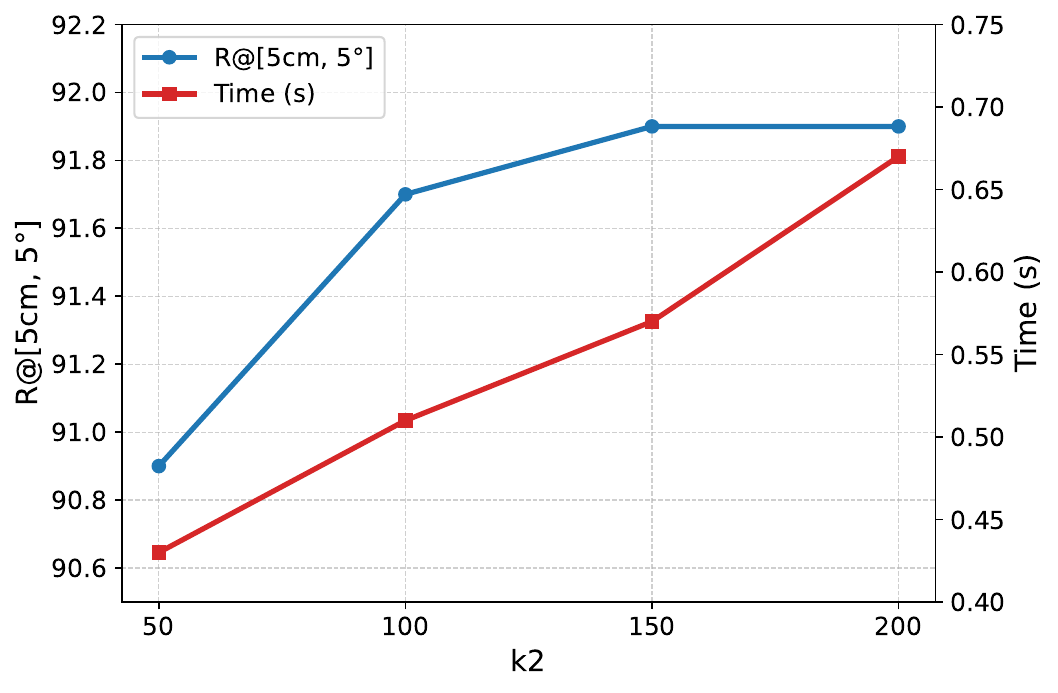}
    \caption{The R@[5cm, 5°] accuracy and the average initial relocalization time under different values of $k_2$.}
    \label{fig: k2_stairs}
    \vspace{-11pt}
\end{figure}

\begin{figure*}[htbp]
    \centering
    \includegraphics[width=0.95\textwidth]{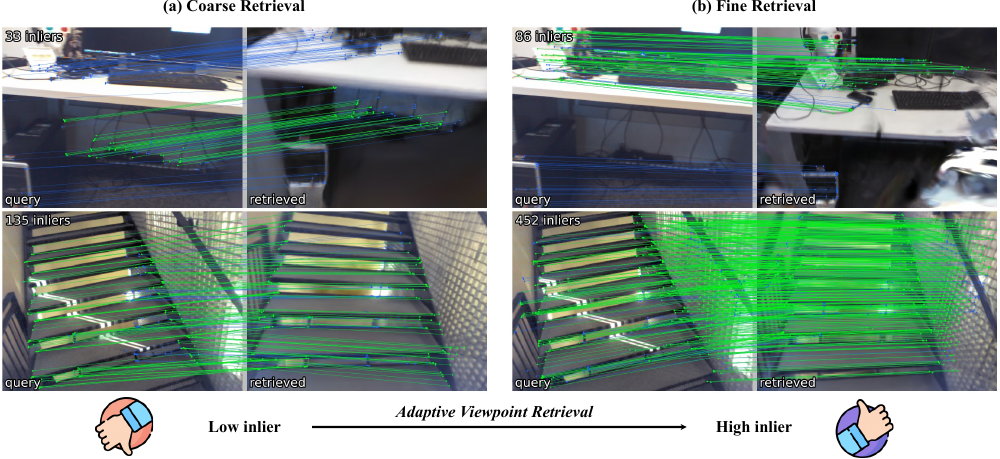}
    \caption{Qualitative visualization of the Adaptive Viewpoint Retrieval process. Green lines denote the inlier correspondences that pass geometric verification, whereas the other matches are visualized in blue. (a) Coarse retrieval often yields candidates with limited co-visibility and low inlier counts. (b) By synthesizing pose-perturbed virtual views and performing fine retrieval, our method significantly increases co-visibility, resulting in substantially more inliers.}
    \vspace{-11pt}
    \label{fig:111}
\end{figure*}

\begin{table*}[t]
    \centering

    \begin{minipage}{0.99\textwidth}
        \renewcommand{\arraystretch}{1.15}
        \centering
        \captionof{table}{The number of images in the training and test sets for each scene.}
        \label{table:Benchmarks}
        \scalebox{0.75}{
        \begin{tabular}{c|c|cccccccccccc}
        \noalign{\hrule height 1pt}
        \multirow{3}{*}{\textbf{7-Scenes}} & \textbf{Scenes} &
        \textbf{\textit{Chess}} & \textbf{\textit{Fire}} & \textbf{\textit{Heads}} &
        \textbf{\textit{Office}} & \textbf{\textit{Pumpkin}} & \textbf{\textit{Kitchen}} &
        \textbf{\textit{Stairs}} & - & - & - & - & - \\
        \cline{2-14}
        & Train & 4000 & 2000 & 1000 & 6000 & 4000 & 7000 & 2000 & - & - & - & - & - \\
        & Test  & 2000 & 2000 & 1000 & 4000 & 2000 & 5000 & 1000 & - & - & - & - & - \\
        \noalign{\hrule height 0.7pt}

        \multirow{3}{*}{\textbf{12-Scenes}} & \textbf{Scenes} &
        \textbf{\textit{Kitchen-1}} & \textbf{\textit{Living-1}} & \textbf{\textit{Bed}} &
        \textbf{\textit{Kitchen-2}} & \textbf{\textit{Living-2}} &
        \textbf{\textit{Luke}} & \textbf{\textit{Gates362}} &
        \textbf{\textit{Gates381}} & \textbf{\textit{Lounge}} &
        \textbf{\textit{Manolis}} & \textbf{\textit{5a}} & \textbf{\textit{5b}} \\
        \cline{2-14}
        & Train & 744 & 1035 & 890 & 782 & 731 & 1370 & 3540 & 2950 & 933 & 1623 & 1000 & 1391 \\
        & Test  & 357 & 493 & 244 & 230 & 359 & 624 & 386 & 1053 & 327 & 807 & 497 & 405 \\
        \noalign{\hrule height 0.7pt}

        \multirow{3}{*}{\textbf{Cambridge}} & \textbf{Scenes} &
        \textbf{\textit{Court}} & \textbf{\textit{College}} &
        \textbf{\textit{Hospital}} & \textbf{\textit{Church}} &
        \textbf{\textit{Shop}} & - & - & - & - & - & - & - \\
        \cline{2-14}
        & Train & 1531 & 1220 & 895 & 1487 & 231 & - & - & - & - & - & - & - \\
        & Test  & 760 & 343 & 182 & 530 & 103 & - & - & - & - & - & - & - \\
        \noalign{\hrule height 1pt}
        \end{tabular}}
    \end{minipage}

    \vspace{10pt}

    \begin{minipage}{0.99\textwidth}
        \renewcommand{\arraystretch}{1.15}
        \centering
        \captionof{table}{The rendered image quality on the test set for each scene.}
        \label{table:PSNR}
        \resizebox{0.99\linewidth}{!}{
        \begin{tabular}{c|c|cccccccccccc}
        \noalign{\hrule height 1pt}
        \multirow{2}{*}{\textbf{7-Scenes}} & \textbf{Scenes} &
        \textbf{\textit{Chess}} & \textbf{\textit{Fire}} & \textbf{\textit{Heads}} &
        \textbf{\textit{Office}} & \textbf{\textit{Pumpkin}} & \textbf{\textit{Kitchen}} &
        \textbf{\textit{Stairs}}& -  & - & - & - & -  \\
        \cline{2-14}
        & PSNR & 24.76 & 22.29 & 18.90 & 22.29 & 24.99 & 21.29 & 19.70 & - & - & - & - & - \\
        \noalign{\hrule height 0.7pt}

        \multirow{2}{*}{\textbf{12-Scenes}} & \textbf{Scenes} &
        \textbf{\textit{Kitchen-1}} & \textbf{\textit{Living-1}} &
        \textbf{\textit{Bed}} & \textbf{\textit{Kitchen-2}} &
        \textbf{\textit{Living-2}} & \textbf{\textit{Luke}} &
        \textbf{\textit{Gates362}} &\textbf{\textit{Gates381}}  &
        \textbf{\textit{Lounge}} &\textbf{\textit{Manolis}} &
        \textbf{\textit{5a}} & \textbf{\textit{5b}}   \\
        \cline{2-14}
        & PSNR & 20.66 & 26.66 & 27.46 & 25.82 & 24.33 & 23.91 &
                 20.63 & 22.82 & 22.85 & 20.31 & 26.81 & 17.72 \\
        \noalign{\hrule height 0.7pt}

        \multirow{2}{*}{\textbf{Cambridge}} & \textbf{Scenes} &
        \textbf{\textit{Court}} & \textbf{\textit{College}} &
        \textbf{\textit{Hospital}} & \textbf{\textit{Church}} &
        \textbf{\textit{Shop}} & - & - & - & - & - & - & -   \\
        \cline{2-14}
        & PSNR & 14.42  & 12.79 & 13.84 & 14.30 & 14.48 & - & - & - & - & - & - & - \\
        \noalign{\hrule height 1pt}
        \end{tabular}}
    \end{minipage}

    \vspace{10pt}

    \begin{minipage}{0.99\textwidth}
        \renewcommand{\arraystretch}{1.8}
        \centering
        \captionof{table}{The relocalization accuracy for each scene on the 12-Scenes dataset.}
        \label{table:12scenes}
        \resizebox{0.99\linewidth}{!}{
        \begin{tabular}{l|cccccccccccc}
        \noalign{\hrule height 1pt}
        \textbf{Scenes} & \textbf{\textit{Kitchen-1}} & \textbf{\textit{Living-1}} &
        \textbf{\textit{Bed}} & \textbf{\textit{Kitchen-2}} &
        \textbf{\textit{Living-2}} & \textbf{\textit{Luke}} &
        \textbf{\textit{Gates362}} &\textbf{\textit{Gates381}}  &
        \textbf{\textit{Lounge}} &\textbf{\textit{Manolis}} &
        \textbf{\textit{5a}} & \textbf{\textit{5b}}   \\
        \cline{1-13}
        \textbf{Avg. Err [cm/°]} $\downarrow$ &
        0.29/0.18 & 0.25/0.12 & 0.24/0.11 & 0.25/0.17 &
        0.25/0.12 & 0.45/0.18 & 0.32/0.13 & 0.30/0.13 &
        0.48/0.14 & 0.32/0.13 & 0.37/0.16 & 0.36/0.14 \\
        \textbf{$R$@[2cm, 2°]} $\uparrow$ &
        98.3 & 100 & 99.6 & 100 & 100 &
        97.3 & 99.7 & 98.8 & 96.9 & 93.1 & 96.6 & 86.9 \\
        \noalign{\hrule height 1pt}
        \end{tabular}}
    \end{minipage}

\end{table*}

\para{Range of perturbations.}
In Fig. \ref{fig:a_stairs}, we vary the angular perturbation $a$ on the Stairs scene.
The accuracy improves as $a$ increases from $0^\circ$ to $5^\circ$, indicating that a moderate amount of angular 
perturbation helps the system escape degenerate initializations and yields better coarse pose 
candidates. However, excessively large perturbations (e.g., $a = 10^\circ$) degrade performance, 
as they place the perturbed poses too far from the true solution.
Fig. \ref{fig:b_stairs} presents the complementary study in which the translation perturbation $b$ is varied on the Stairs scene. Introducing a small positional perturbation (e.g., $b=0.2$\,m) already produces a noticeable improvement, and the performance 
peaks at moderate magnitudes ($b = 0.5{\text{--}}1.0$\,m). When the perturbation becomes too 
large (e.g., $b=1.5$\,m), the accuracy drops, as the perturbed viewpoints deviate excessively 
from plausible poses and reduce the reliability of the rendered feature cues.
Based on these observations, we adopt $(a = 5^\circ, b = 0.5\,\text{m})$ as the default configuration in the adaptive retrieval pipeline.

\para{Number of perturbations.} A larger $k_2$ provides more pose-perturbed virtual views, increasing the chance of finding a viewpoint closer to the query.
As shown in Fig.~\ref{fig: k2_stairs}, the recall R@[5cm, 5°] on the Stairs scene improves rapidly when $k_2$ increases from 50 to 100, and marginally from 100 to 150, while no further gain is observed at $k_2 = 200$.
In contrast, the initialization time grows approximately linearly with $k_2$, since more virtual views are rendered and verified.
Considering this trade-off between accuracy and efficiency, we adopt $k_2 = 150$ for indoor scene.


\para{Visualization.}
Fig.~\ref{fig:111} provides a qualitative visualization of the Adaptive Viewpoint Retrieval process.
The coarse retrieval stage selects reference images with limited viewpoint overlap due to the sparse image observations, resulting in only a small number of geometrically consistent matches.
After synthesizing pose-perturbed virtual views, the fine retrieval stage identifies reference images that are significantly closer to the query viewpoint.
This leads to a substantial increase in co-visible regions and a much larger set of verified inliers, as illustrated by the dense green correspondences on the right.

\section{Scene Metadata}\label{app:scene_meta}

We report the number of training and test images for each scene in the 7-Scenes, 12-Scenes, and Cambridge Landmarks datasets in Table~\ref{table:Benchmarks}.
Moreover, we apply sky and dynamic-object masks to exclude these regions from supervision, preventing their instability from degrading the FGS training process in the Cambridge Landmarks dataset.

\section{Mapping Quality}\label{app:mapping_quality}

We report the PSNR of the FGS models trained within our SplatHLoc framework for each scene in Table~\ref{table:PSNR}. 
In indoor scenes, the PSNR typically reaches around 20 dB, while in outdoor scenes it is closer to 15 dB.
This trend is consistent with the characteristics of the datasets: indoor environments contain more stable lighting conditions and richer geometric structures, enabling the FGS model to fit the scene appearance more accurately. 
In contrast, outdoor scenes exhibit stronger illumination changes, dynamic elements, and larger viewpoint variations, making high-fidelity rendering more challenging. Moreover, the Cambridge Landmarks dataset provides fewer training images for each scene.
Nevertheless, the obtained PSNR values demonstrate that the trained FGS maps retain sufficient photometric fidelity to support robust retrieval and matching within our pipeline.

\section{Additional Experimental Results}\label{app:extra_exp}

\para{12-Scenes.} We report the relocalization results of our SplatHLoc on each scene of the 12-Scenes dataset in Table \ref{table:12scenes}.
Across all scenes, our method achieves low translation and rotation errors, with most scenes exhibiting average errors below 0.3 cm and 0.15°.
The corresponding recall under the stringent R@[2 cm, 2°] metric reaches nearly 100\% on the majority of scenes, demonstrating the robustness of SplatHLoc in challenging indoor environments.

\begin{figure}[htbp]
    \centering
    \includegraphics[width=0.45\textwidth]{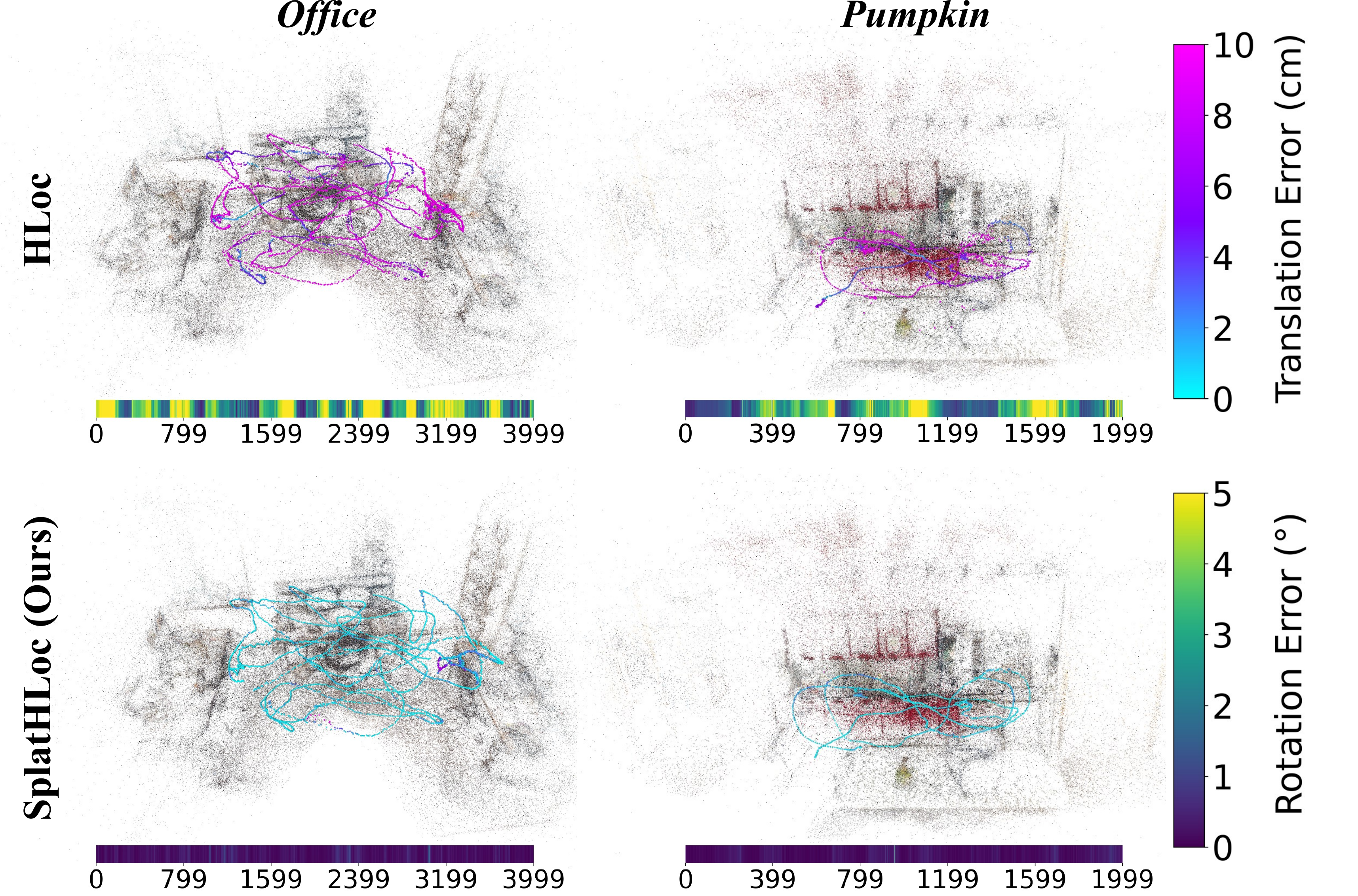}
    \caption{Additional comparison of camera pose estimation errors for the 7-Scenes dataset.}
    \vspace{-11pt}
    \label{fig:vis_error}
\end{figure}

\para{Map size.} Our method requires storing additional VPR features of database images during the mapping process. In Table~\ref{map} (main text), we only compared the size of the FGS map. For the chess scene, storing the VPR features requires 62.50 MB. Nonetheless, the total map size of our method for the chess scene remains less than half that of STDLoc.

\begin{figure}[htbp]
    \centering
    \includegraphics[width=0.46\textwidth]{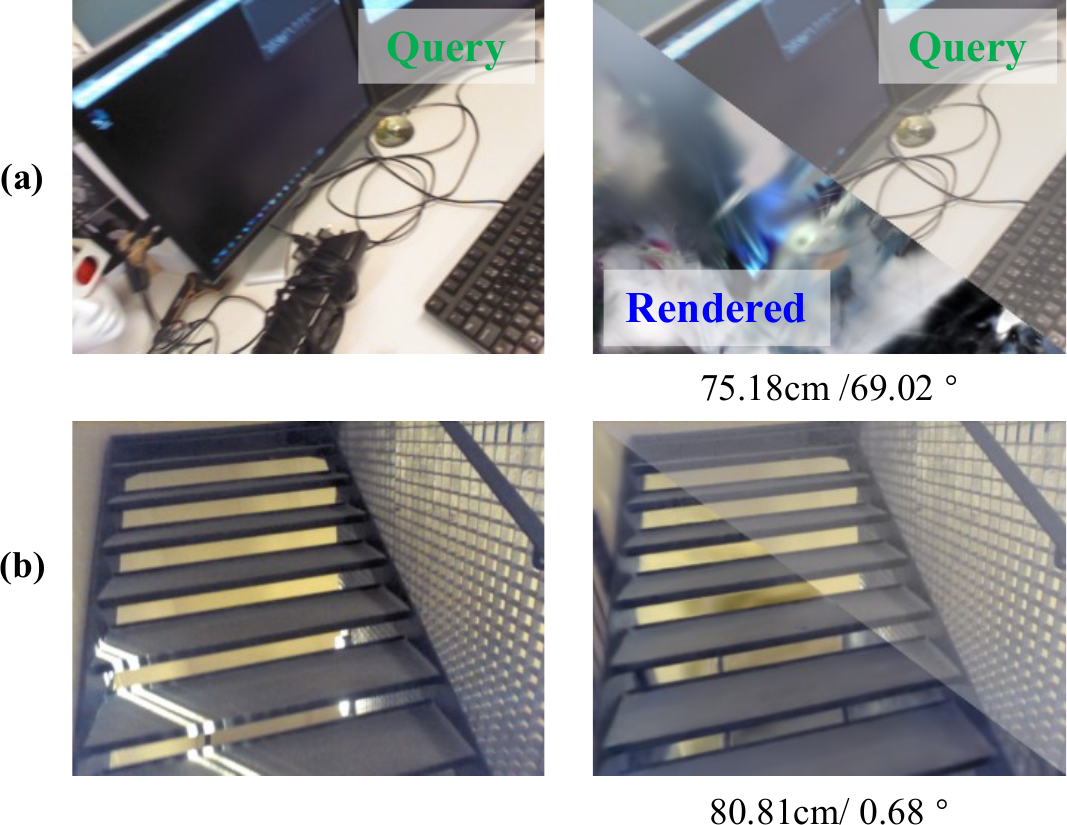}
    \caption{Visualization of the failure cases. (a) The first row illustrates a failure case caused by poor rendering quality of the Gaussian map near sparsely observed viewpoints. (b) The second row shows a failure induced by repetitive structures in the scene.}
    \vspace{-11pt}
    \label{fig:fail_cases}
\end{figure}

\section{Additional Visualization}\label{app:extra_vis}

\para{Relocalization error.}
Fig.~\ref{fig:vis_error} shows a comparison of the camera pose estimation errors in the \textit{Office} and \textit{Pumpkin} scenes.
The error maps further show that SplatHLoc reduces outlier predictions and maintains lower errors across the entire sequence, demonstrating superior robustness in challenging indoor scenes.
For trajectory points exceeding the upper error limit (10cm, 2°), the color is set to the value corresponding to that limit.

\begin{figure*}[htbp]
    \centering
    \includegraphics[width=0.97\textwidth]{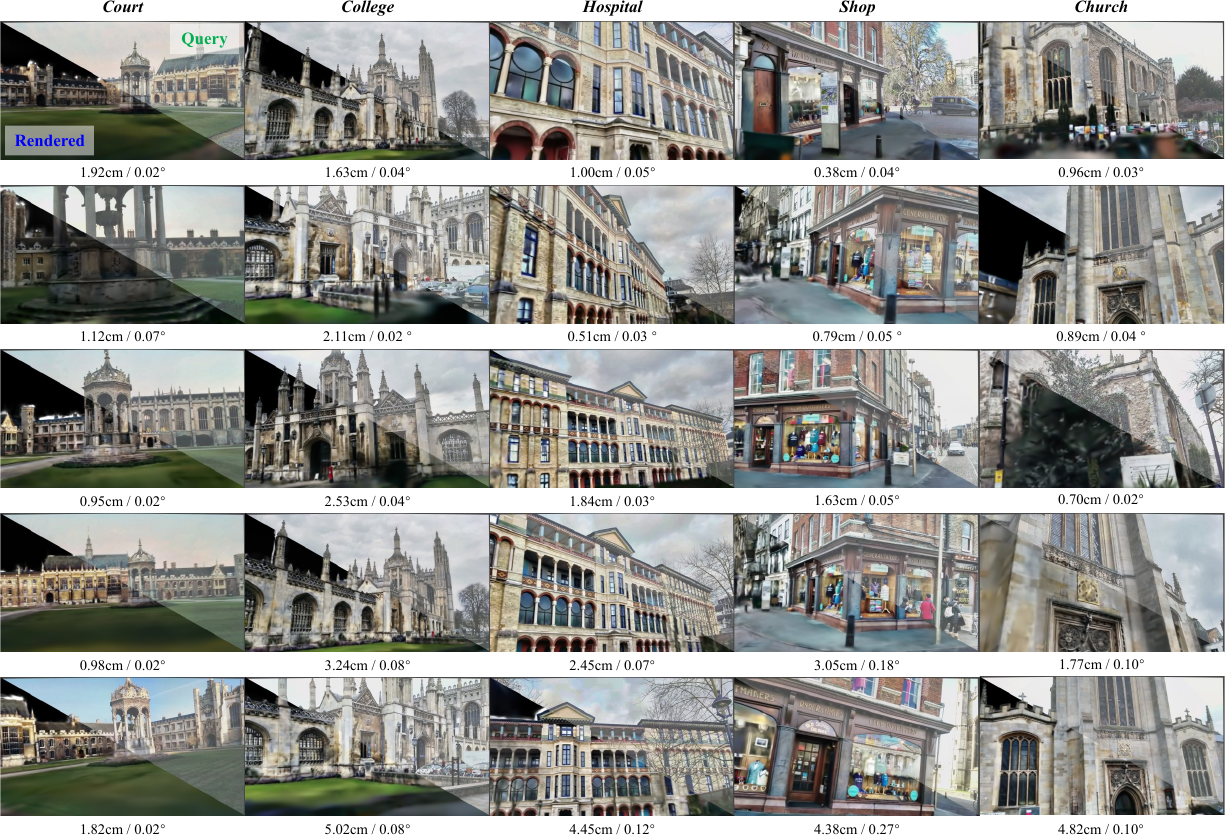}
    \caption{Qualitative visualization of the relocalization errors on the Cambridge Landmarks dataset. For each image pair, the upper-left triangle shows the query image, while the lower-right triangle displays the rendered image obtained from the final estimated pose. Better visual alignment between the two indicates that our SplatHLoc achieves more accurate relocalization.}
    \vspace{-11pt}
    \label{fig:cam_error}
\end{figure*}

Fig.~\ref{fig:cam_error} provides qualitative visualizations of the relocalization errors across all scenes in the Cambridge Landmarks dataset.
Across Court, College, Hospital, Shop, and Church, the rendered views closely align with the corresponding query images, demonstrating that SplatHLoc achieves highly accurate pose estimation even in large-scale outdoor environments.

\begin{figure*}[htbp]
    \centering
    \includegraphics[width=0.97\textwidth]{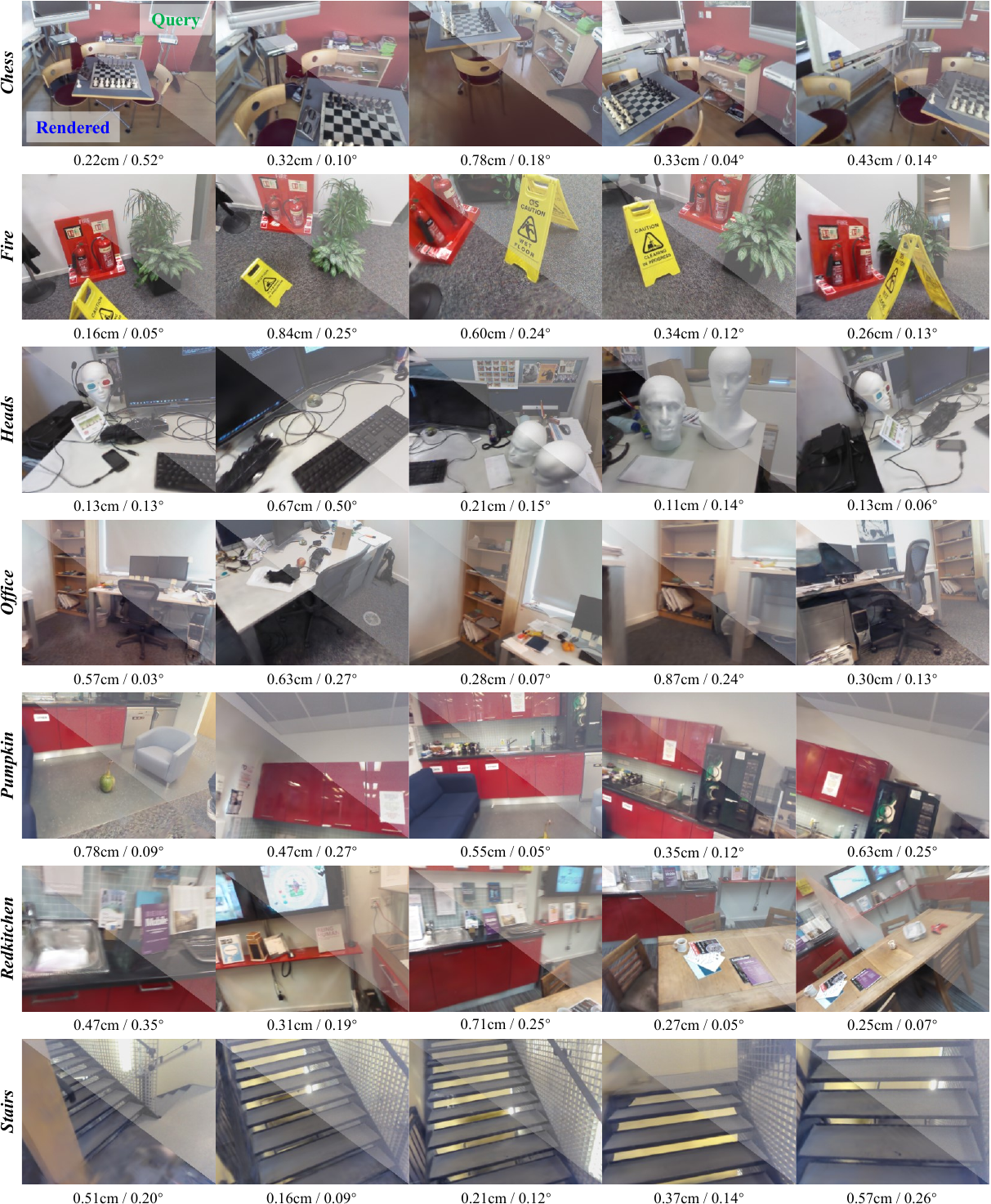}
    \caption{Qualitative visualization of the relocalization errors on the 7-Scenes dataset. For each image pair, the upper-left triangle shows the query image, while the lower-right triangle displays the rendered image obtained from the final estimated pose. Better visual alignment between the two indicates that our SplatHLoc achieves more accurate relocalization.}
    \vspace{-11pt}
    \label{fig:7s_error}
\end{figure*}

In Fig.~\ref{fig:7s_error}, we visualize the relocalization errors for each scene in the 7-Scenes dataset.
Across all scenes, SplatHLoc consistently produces well-aligned query–rendered pairs, demonstrating the robustness and reliability of our approach under varying levels of texture, clutter, and viewpoint change.

\para{Failure Cases.}
Fig.~\ref{fig:fail_cases} presents representative failure cases observed during evaluation.
In case (a), the failure arises from poor rendering quality near sparsely observed viewpoints in the training set.
When the Gaussian map does not sufficiently cover a region, the synthesized image deviates significantly from the true appearance, leading to ultimately incorrect pose estimation.
Case (b) shows a different type of failure induced by highly repetitive structures in the scene.
In such environments, distinct locations may produce visually similar local patterns, causing the feature matching stage to converge to an incorrect alignment despite low rotational ambiguity.
These cases highlight challenging scenarios where even high-quality novel view synthesis may be insufficient for accurate relocalization.

\section{Gaussian Rasterization in FGS}\label{app:fgs_raster}

We adopt the differentiable rasterization procedure of 3D Gaussian Splatting.
Each Gaussian primitive is parameterized as  
$\mathcal{G}_i = \{\mathbf{x}_i,\mathbf{q}_i,\mathbf{s}_i,\alpha_i,\mathbf{c}_i,\mathbf{f}_i\}$,  
where $\mathbf{x}_i\in\mathbb{R}^3$ denotes the center,  
$\mathbf{q}_i$ the rotation,  
$\mathbf{s}_i$ the anisotropic scale,  
$\alpha_i$ the base opacity,  
$\mathbf{c}_i$ the color,  
and $\mathbf{f}_i\in\mathbb{R}^d$ the feature vector.

\vspace{4pt}
\para{3D covariance in world coordinates.}
The intrinsic 3D covariance of Gaussian $i$ in the world coordinate frame is
\begin{equation}
\Sigma^{3D}_{i,\mathrm{world}}
= R(\mathbf{q}_i)\,
\mathrm{diag}(\mathbf{s}_i^2)\,
R(\mathbf{q}_i)^\top,
\end{equation}
where $R(\mathbf{q}_i)$ is the rotation matrix associated with quaternion~$\mathbf{q}_i$.

\vspace{4pt}
\para{Transformation to the camera frame.}
Given the viewing transformation $W \in \mathrm{SO}(3)$ (the rotation of world-to-camera pose),
the 3D covariance in the camera coordinate system becomes
\begin{equation}
\Sigma^{3D}_{i,\mathrm{cam}}
= W\,\Sigma^{3D}_{i,\mathrm{world}}\,W^\top.
\end{equation}

\vspace{4pt}
\para{Projection to the image plane.}
Under the camera projection $\pi:\mathbb{R}^3 \rightarrow \mathbb{R}^2$,  
the corresponding 2D covariance is obtained by linearizing $\pi$ at the Gaussian center:
\begin{equation}
\Sigma^{2D}_i
= J_i\,\Sigma^{3D}_{i,\mathrm{cam}}\,J_i^\top,
\qquad
J_i = \left.\frac{\partial \pi(\mathbf{x})}{\partial \mathbf{x}}
\right|_{\mathbf{x}=\mathbf{x}_i}.
\end{equation}

\vspace{4pt}
\para{Projected kernel and opacity.}
The 2D Gaussian kernel on the image plane is
\begin{equation}
G_i(\mathbf{u})
= \exp\!\left(
-\tfrac{1}{2}
(\mathbf{u}-\mu_i)^\top (\Sigma^{2D}_i)^{-1}(\mathbf{u}-\mu_i)
\right),
\end{equation}
where $\mu_i = \pi(\mathbf{x}_i)$.
The per-pixel opacity is
\begin{equation}
\alpha_i(\mathbf{u}) = \alpha_i\, G_i(\mathbf{u}).
\end{equation}

\vspace{4pt}
\para{Volume compositing.}
Given Gaussians sorted in front-to-back order, the accumulated transmittance is
\begin{equation}
T_i = \prod_{j < i} (1 - \alpha_j).
\end{equation}
The rendered color is obtained via alpha compositing:
\begin{equation}
\mathbf{C}(\mathbf{u})
= \sum_{i \in \mathcal{N}(\mathbf{u})}
\mathbf{c}_i\,
\alpha_i(\mathbf{u})\,
T_i(\mathbf{u}),
\end{equation}
where $\mathcal{N}(\mathbf{u})$ denotes the Gaussians influencing pixel $\mathbf{u}$.

\vspace{4pt}
\para{Depth and feature rendering.}
Depth rendering replaces the color with the camera-space depth $z_i$:
\begin{equation}
\mathbf{D}(\mathbf{u})
= \sum_{i \in \mathcal{N}(\mathbf{u})}
z_i\,
\alpha_i(\mathbf{u})\,
T_i(\mathbf{u}).
\end{equation}
Feature rendering follows the similar form:
\begin{equation}
\mathbf{F}(\mathbf{u})
= \mathrm{norm}\!\left(
\sum_{i \in \mathcal{N}(\mathbf{u})}
\mathrm{norm}(\mathbf{f}_i)\,
\alpha_i(\mathbf{u})\,
T_i(\mathbf{u})
\right),
\end{equation}
where $\mathrm{norm}(\cdot)$ denotes the L2 normalization operation.

\vspace{4pt}
\para{Differentiability.}
All operations—covariance transformation, projection, kernel evaluation,
opacity accumulation, and compositing—are differentiable with respect to  
$\{\mathbf{x}_i,\mathbf{q}_i,\mathbf{s}_i,\alpha_i,\mathbf{c}_i,\mathbf{f}_i\}$,
enabling the end-to-end optimization of all Gaussian attributes.

\begin{algorithm*}[t]
\caption{Adaptive Coarse-to-Fine Viewpoint Retrieval}
\label{alg:adaptive_c2f_retrieval}
\begin{algorithmic}[1]
\REQUIRE Query image $I_q$, Gaussian map $\mathcal{G}$, 
         VPR model $\mathcal{V}$, sparse matcher $\mathcal{M}_{\text{sparse}}$, \\
         training images $\{I_t\}$ with poses $\{\mathbf{T}_t\}$, \\
         retrieval hyperparameters $(k_1, k_2, k_3, a, b, \mathcal{I})$.
\ENSURE Final candidate image $I_c$ and its pose $\mathbf{T}^\star$ for $I_q$.

\STATE \textbf{// Coarse viewpoint retrieval}
\STATE Extract global descriptor $\mathbf{v}_q = \mathcal{V}(I_q)$.
\STATE Compute descriptors $\{\mathbf{v}_t\}$ for all training images.
\STATE Retrieve top-$k_1$ neighbors $\{I_c^1, \dots, I_c^{k_1}\}$ of $I_q$.
\STATE Initialize $N^\star \leftarrow 0$, \; $\mathbf{T}^\star \leftarrow \varnothing$, \; $i^\star \leftarrow -1$.

\FOR{$i = 1$ to $k_1$}
    \STATE $I_c \leftarrow I_c^i$; \quad $\mathbf{T}_c \leftarrow$ pose of $I_c$ from $\{\mathbf{T}_t\}$.
    \STATE $\text{matches} \leftarrow \mathcal{M}_{\text{sparse}}(I_q, I_c)$.
    \STATE Estimate inliers $\mathcal{I}_c$.
    \STATE $N \leftarrow |\mathcal{I}_c|$.
    \IF{$N > N^\star$}
        \STATE $N^\star \leftarrow N$, \quad $\mathbf{T}^\star \leftarrow \hat{\mathbf{T}}$, \quad $i^\star \leftarrow i$.
    \ENDIF
    \IF{$N^\star \geq \mathcal{I}$}
        \STATE \textbf{break} \quad \text{// sufficient inliers, stop coarse search}
    \ENDIF
\ENDFOR
\STATE $I_c^c \leftarrow I_c^{i^\star}$ \quad \text{// best coarse candidate image}

\STATE \textbf{// Fine retrieval via virtual novel viewpoints}
\STATE $I_c^f \leftarrow I_c^c$ \quad \text{// by default, fall back to the coarse candidate}
\IF{$N^\star < \mathcal{I}$}
    \STATE Generate $k_2$ perturbed poses $\{\mathbf{T}_v^j\}_{j=1}^{k_2}$ around $\mathbf{T}^{\star}$ within $(a^\circ,\, b\text{ m})$
.
    \FOR{$j = 1$ to $k_2$}
        \STATE Render virtual view $I_v^j = \text{Render}(\mathcal{G}, \mathbf{T}_v^j)$.
        \STATE Extract descriptor $\mathbf{v}_v^j = \mathcal{V}(I_v^j)$.
    \ENDFOR
    \STATE Build a VPR database with $\{\mathbf{v}_v^j\}_{j=1}^{k_2}$.
    \STATE Retrieve top-$k_3$ virtual views $\{I_v^{j_1}, \dots, I_v^{j_{k_3}}\}$ for $I_q$.
    \STATE Initialize $j^\star \leftarrow -1$.
    \FOR{each index $j \in \{j_1, \dots, j_{k_3}\}$}
        \STATE $\text{matches} \leftarrow \mathcal{M}_{\text{sparse}}(I_q, I_v^j)$.
        \STATE Estimate pose inliers $\mathcal{I}_v$.
        \STATE $N \leftarrow |\mathcal{I}_v|$.
        \IF{$N > N^\star$}
            \STATE $N^\star \leftarrow N$, \; $\mathbf{T}^\star \leftarrow \hat{\mathbf{T}}$, \; $j^\star \leftarrow j$.
        \ENDIF
    \ENDFOR
    \IF{$j^\star \ge 0$}
        \STATE $I_c^f \leftarrow I_v^{j^\star}$ \quad \text{// best virtual candidate image}
    \ENDIF
\ENDIF

\STATE $I_c \leftarrow I_c^f$ \quad \text{// final candidate image}
\RETURN $I_c, \mathbf{T}^\star$.
\end{algorithmic}
\vspace{30pt}
\end{algorithm*}

\end{document}